\documentclass{article}

\PassOptionsToPackage{numbers, compress}{natbib}

\usepackage[final]{neurips_2021}

\usepackage[utf8]{inputenc} %
\usepackage[T1]{fontenc}    %
\usepackage{hyperref}       %
\usepackage{url}            %
\usepackage{booktabs}       %
\usepackage{amsfonts}       %
\usepackage{nicefrac}       %
\usepackage{microtype}      %
\usepackage{xcolor}         %
\usepackage{amsmath}
\usepackage{amssymb}
\usepackage{bm}
\usepackage{textcomp}
\usepackage{graphicx}
\usepackage{subfig}
\usepackage{tikz}
\usetikzlibrary{shapes.geometric}
\usetikzlibrary{patterns}
\usepackage{bbm}
\usepackage{wrapfig}
\usepackage{algorithmic}
\usepackage{algorithm}
\usepackage{multirow}
\usepackage{todonotes}
\usepackage{enumitem}
\usepackage{pifont}
\newcommand{\cmark}{\ding{51}}%
\newcommand{\xmark}{\ding{55}}%

\newcommand{\dmin}{d_{\mathrm{min}}}
\newcommand{\dmax}{d_{\mathrm{max}}}

\newcommand{\bR}{\mathbb{R}}

\newcommand{\vA}{\mathbf{A}}
\newcommand{\vP}{\mathbf{P}}

\newcommand{\vx}{\mathbf{x}}
\newcommand{\vu}{\mathbf{u}}
\newcommand{\vy}{\mathbf{y}}
\newcommand{\vv}{\mathbf{v}}

\newcommand{\vxt}[1]{\mathbf{x}_{#1}}
\newcommand{\vyt}[1]{\mathbf{y}_{#1}}

\newcommand{\electricity}{\texttt{electricity}}
\newcommand{\traffic}{\texttt{traffic}}
\newcommand{\solar}{\texttt{solar}}
\newcommand{\wiki}{\texttt{wiki}}
\newcommand{\exchange}{\texttt{exchange}}
\newcommand{\bb}{\texttt{bouncing ball}}
\newcommand{\edslds}{\texttt{3 mode system}}
\newcommand{\bee}{\texttt{dancing bees}}

\newcommand{\DeepAR}{\texttt{DeepAR}}
\newcommand{\DeepState}{\texttt{DeepState}}

\newcommand{\RSGLSISSM}{\texttt{RSGLS-ISSM}}
\newcommand{\ARSGLS}{\texttt{ARSGLS}}
\newcommand{\KVAE}{\texttt{KVAE}}
\newcommand{\KVAEMC}{\texttt{KVAE-MC}}
\newcommand{\KVAERB}{\texttt{KVAE-RB}}
\newcommand{\SNLDS}{\texttt{SNLDS}}
\newcommand{\REDSDS}{\texttt{RED-SDS}}
\newcommand{\EDSDS}{\texttt{ED-SDS}}

\graphicspath{ {images/} }
\title{Deep Explicit Duration Switching Models\\for Time Series}

\makeatletter
\newcommand{\specificthanks}[1]{\@fnsymbol{#1}}%
\makeatother

\author{%
  Abdul Fatir Ansari{\normalfont\textsuperscript{2}}\thanks{Equal contribution.}$\enspace$\thanks{Work done during an internship at Amazon Research.}
  \quad Konstantinos Benidis{\normalfont\textsuperscript{1}}\footnotemark[1]
  \quad Richard Kurle{\normalfont\textsuperscript{1}}
  \quad Ali Caner T\"{u}rkmen{\normalfont\textsuperscript{1}}\\[10pt]
  {\bfseries
  \quad Harold Soh{\normalfont\textsuperscript{2}}
  \quad Alexander J. Smola{\normalfont\textsuperscript{1}}
  \quad Yuyang Wang{\normalfont\textsuperscript{1}}
  \quad Tim Januschowski{\normalfont\textsuperscript{1}}
  }\\[10pt]
  \textsuperscript{1}Amazon Research
  \quad
  \textsuperscript{2}National University of Singapore\\[6pt]
  Correspondence to: \texttt{abdulfatir@u.nus.edu}, \texttt{\{kbenidis, kurler\}@amazon.com}
}

\begin{document}

\maketitle

\begin{abstract}
Many complex time series can be effectively subdivided into distinct regimes that exhibit persistent dynamics. Discovering the switching behavior and the statistical patterns in these regimes is important for understanding the underlying dynamical system. 
We propose the Recurrent Explicit Duration Switching Dynamical System (RED-SDS), a flexible model that is capable of identifying both state- and time-dependent switching dynamics. 
State-dependent switching is enabled by a recurrent state-to-switch connection and an explicit duration count variable is used to improve the time-dependent switching behavior. 
We demonstrate how to perform efficient inference using a hybrid algorithm that approximates the posterior of the continuous states via an inference network and performs exact inference for the discrete switches and counts.
The model is trained by maximizing a Monte Carlo lower bound of the marginal log-likelihood that can be computed efficiently as a byproduct of the inference routine.
Empirical results on multiple datasets demonstrate that RED-SDS achieves  considerable improvement in time series segmentation and competitive forecasting performance against the state of the art. 
\end{abstract}

\section{Introduction}
\label{sec:intro}

Time series forecasting plays a key role in informing industrial and business decisions~\cite{petropoulos2020forecasting,tim2019class, benidis2020neural}, while segmentation is useful for understanding biological and physical systems~\cite{oh2008learning,sharma2018point,linderman2019hierarchical}.
State Space Models (SSMs) \cite{durbin2012time} are a powerful tool for such tasks---especially when combined with neural networks \cite{rangapuram2018deep, de2020normalizing,dong2020collapsed}---since they provide a principled framework for time series modeling. 
One of the most popular SSMs is the Linear Dynamical System (LDS)~\cite{bar1993estimation,roweis1999unifying}, which models the dynamics of the data using a continuous latent variable, called \emph{state}, that evolves with Markovian linear transitions. 
The assumptions of LDS allow for exact inference of the states~\cite{kalman1960new};  however, they are too restrictive for real-world systems that often exhibit piecewise linear or non-linear hidden dynamics with a finite number of operating modes or \emph{regimes}. 
For example, the power consumption of a city may follow different hidden dynamics during weekdays and weekends. 
Such data are better explained by a Switching Dynamical System (SDS)~\cite{ackerson1970state,ghahramani2000variational}, an SSM with an additional set of latent variables called \emph{switches} that define the operating mode active at the current timestep. 

Switching events can be classified into time-dependent or state-dependent~\cite{liberzon2003switching}. Historically, emphasis was placed on the former, which occurs after a certain amount of time has elapsed in a given regime. While in a vanilla SDS switch durations follow a geometric distribution, more complex long-term temporal patterns can be captured using explicit duration models~\cite{oh2008learning,chiappa2019explicit}. As a recent alternative to time-dependency, recurrent state-to-switch connections~\cite{linderman2016recurrent} have been proposed that capture state-dependent switching, i.e., a change that occurs when the state variable enters a region that is governed by a different regime. For added flexibility, these models can be used in conjunction with transition/emission distributions parameterized by neural networks~\cite{Johnson2016ComposingGM,fraccaro2017disentangled,dong2020collapsed,kurle2020deeprao}. Recent works, e.g.,  \cite{dong2020collapsed, kurle2020deeprao}, proposed hybrid inference algorithms that exploit the graphical model structure to perform approximate inference for some latent variables and conditionally exact inference for others. 

\begin{wrapfigure}[13]{r}{0.63\textwidth}
\vspace{-2mm}
	\includegraphics[width=0.99\linewidth]{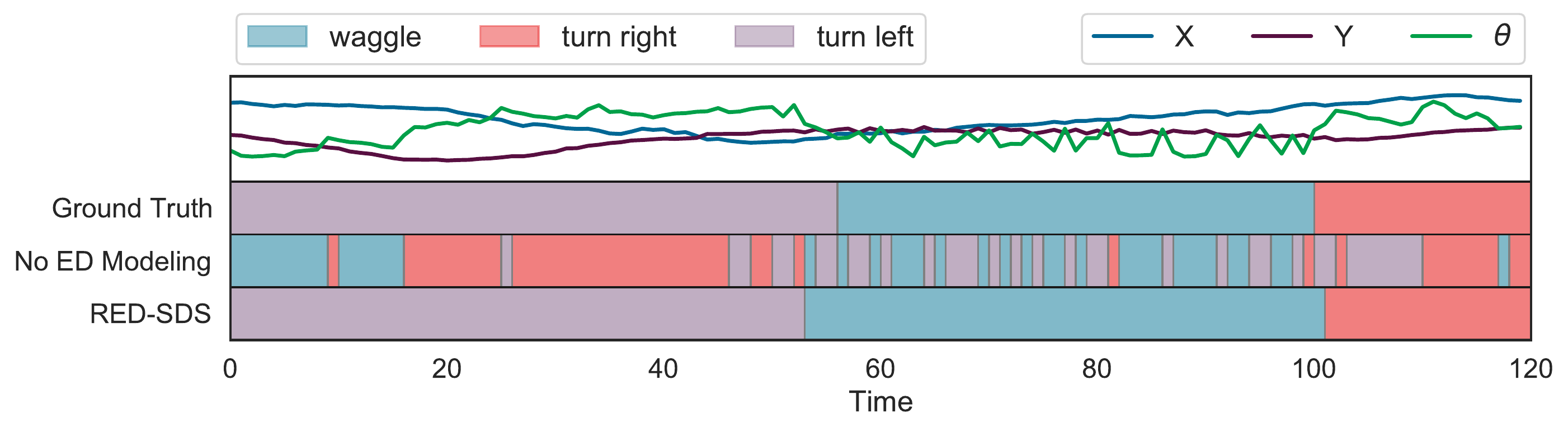}
	\caption{\small Segments (colored bars at the bottom) inferred by a baseline with no Explicit Duration (ED) modeling vs. our RED-SDS for a time series from the dancing bees dataset (top). The baseline struggles to learn long-term temporal patterns, particularly during the ``waggle'' phase of the bee dance.}
	\label{fig:main-fig}
\end{wrapfigure}
Despite these advances in representation and inference, modeling complex real-world temporal phenomena remains challenging. For example, state-of-the-art state-dependent models (e.g., \cite{dong2020collapsed}) lack the capacity to adequately capture time-dependent switching. Empirically, we find this hampers their ability to learn parsimonious segmentations when faced with complex patterns and long-term dependencies (see Fig. \ref{fig:main-fig} for an example). Conversely, time-dependent switching models are ``open-loop'' and unable to model state-conditional behavioral transitions that are common in many systems, e.g., in autonomous or multi-agent systems~\cite{linderman2016recurrent}. Intuitively, the suitability of the switching model largely depends on the underlying data-generating process; city power consumption may be better modeled via time-dependent switching, whilst the motion of a ball bouncing between two walls is driven by its state. Indeed, complex real-world processes likely involve both types of switching behavior.

Motivated by this gap, we propose the Recurrent Explicit Duration Switching Dynamical System (RED-SDS) that captures \emph{both} state-dependent and time-dependent switching. RED-SDS combines the recurrent state-to-switch connection with explicit duration models for switches. %
Notably, RED-SDS allows the incorporation of inductive biases via the hyperparameters of the duration models to better identify long-term temporal patterns. However, this combination also complicates inference, especially when using neural networks to model the underlying probability distributions. To address this technical challenge, we propose a hybrid algorithm that (i) uses an inference network for the continuous latent variables (states) and (ii) performs efficient exact inference for the discrete latent variables (switches and counts) using a forward-backward routine similar to Hidden Semi-Markov Models~\cite{yu2010hidden,chiappa2019explicit}. The model is trained by maximizing a Monte Carlo lower bound of the marginal log-likelihood that can be efficiently computed by the inference routine.

We evaluated RED-SDS on two important tasks: segmentation and forecasting. Empirical results on segmentation show that RED-SDS is able to identify both state- and time-dependent switching patterns, considerably outperforming benchmark models. 
For example, Fig. \ref{fig:main-fig} shows that RED-SDS addresses the oversegmentation that occurs with an existing strong baseline~\cite{dong2020collapsed}. 
For forecasting, we illustrate the competitive performance of RED-SDS with an extensive evaluation against state-of-the-art models on multiple benchmark datasets. 
Further, we show how our model is able to simplify the forecasting problem by breaking the time series into different meaningful regimes without any imposed structure. As such, we manage to learn appropriate duration models for each regime and extrapolate the learned patterns into the forecast horizon consistently.   

In summary, the key contributions of this paper are:
\begin{itemize}
\item  RED-SDS, a novel non-linear state space model which combines the recurrent state-to-switch connection with explicit duration models to flexibly model switch durations;
\item an efficient hybrid inference and learning algorithm that combines approximate inference for states with conditionally exact inference for switches and counts;
\item a thorough evaluation on a number of benchmark datasets for time series segmentation and forecasting, demonstrating that RED-SDS can learn meaningful duration models, identify both state- and time-dependent switching patterns and extrapolate the learned patterns consistently into the future.
\end{itemize}

\section{Background: switching dynamical systems}
\label{sec:background}

\paragraph{Notation.} Matrices, vectors and scalars are denoted by uppercase bold, lowercase bold and lowercase normal letters, respectively. We denote the sequence $\{\vyt{1}, \dots, \vyt{T}\}$ by $\vyt{1:T}$, where $\vyt{t}$ is the value of $\vy$ at time $t$. In our notation, we do not further differentiate between random variables and their realizations.

Switching Dynamical Systems (SDS) are hybrid SSMs that use discrete ``switching'' states $z_t$ to index one of $K$ base dynamical systems with continuous states $\vxt{t}$. 
The joint distribution factorizes as 
\begin{equation}
p(\vyt{1:T}, \vxt{1:T}, z_{1:T}) = \prod_{t=1}^{T}  p(\vyt{t} | \vxt{t}) p(\vxt{t}|\vxt{t-1}, z_t) p(z_t | z_{t-1}),
\end{equation}
where $p(\vxt{1}|\vxt{0}, z_1) p(z_1|z_0) = p(\vxt{1} | z_1) p(z_1)$ is the initial (continuous and discrete) state prior. 
The base dynamical systems have continuous state transition $p(\vxt{t} | \vxt{t-1}, z_t)$ and continuous or discrete emission $p(\vyt{t} | \vxt{t})$ that can both be linear or non-linear.

The discrete transition $p(z_t | z_{t-1})$ of vanilla SDS is parametrized by a stochastic transition matrix $\mathbf{A} \in \mathbb{R}^{K \times K}$, where the entry $a_{ij} = \mathbf{A}(i, j)$ represents the probability of switching from state $i$ to state $j$. This results in an ``open loop'' as the transition only depends on the previous switch which inhibits the model from learning state-dependent switching patterns~\cite{linderman2016recurrent}. Further, the state duration (also known as the \emph{sojourn time}) follows a geometric distribution~\cite{chiappa2019explicit}, where the probability of staying in state $i$ for $d$ steps is $\rho_i(d) = (1 - a_{ii})a_{ii}^{d-1}$. 
This {\em memoryless} switching process results in frequent regime switching, limiting the ability to capture consistent long-term time-dependent switching patterns.
In the following, we briefly discuss two approaches that have been proposed to improve the state-dependent %
and time-dependent switching %
capabilities in SDSs.

\textbf{Recurrent SDS.} 
Recurrent SDSs (e.g.,~\cite{barber2006ecslds, linderman2016recurrent, becker2019sdvbf, kurle2020deeprao}) address state-dependent switching by changing the switch transition distribution to $p(z_t | \vx_{t-1}, z_{t-1})$---called the state-to-switch recurrence---implying that the switch transition distribution changes at every step and the sojourn time no longer follows a geometric distribution. 
This extension complicates inference.
Furthermore, the first-order Markovian recurrence does not adequately address long-term time-dependent switching.

\textbf{Explicit duration SDS.} 
Explicit duration SDSs are a family of models that introduce additional random variables to explicitly model the switch duration distribution. 
Explicit duration variables have been applied to both HMMs and SDSs with Gaussian linear continuous states; the resulting models are referred to as Hidden Semi-Markov Models (HSMMs) \cite{murphy2002hidden,yu2010hidden}, and Explicit Duration Switching Linear Gaussian SSMs (ED-SLGSSMs) \cite{chiappa2019explicit,oh2008learning,chiappa2010movement}, respectively.
Several methods have been proposed in the literature for modeling the switch duration, e.g., using decreasing or increasing count, and duration-indicator variables.
In the following, we briefly describe modeling switch duration using increasing count variables and refer the reader to \citet{chiappa2019explicit} for details. 

Increasing count random variables $c_t$ represent the \emph{run-length} of the currently active regime and can either increment by 1 or reset to 1.
An increment indicates that the switch variable $z_t$ is copied over to the next timestep whereas a reset indicates a regular Markov transition using the transition matrix $\mathbf{A}$.
Each of the $K$ switches has a distinct duration distribution $\rho_k$, a categorical distribution over $\{\dmin, \dots, d_{\mathrm{max}}\}$, where $\dmin$ and $\dmax$ delimit the number of steps before making a Markov transition. 
Following~\cite{oh2008learning,chiappa2019explicit}, the probability of a count increment is given by
\begin{equation}
\label{eq:count_incr}
	v_k(c) = 1 - \frac{\rho_k(c)}{\sum^{d_{\mathrm{max}}}_{d=c}\rho_k(d)}.
\end{equation} 
The transition of count $c_t$ and switch $z_t$ variables is defined as
\begin{align}
	p(c_t | z_{t-1} = k, c_{t-1} ) &= 
	\begin{cases}
		v_{k}(c_{t-1}) & \text{if}\quad c_t = c_{t-1} + 1\\
		1 - v_{k}(c_{t-1}) & \text{if}\quad c_t = 1\\
	\end{cases},\\
	p(z_t = j | z_{t-1} = i, c_t) &= 
	\begin{cases}
		\delta_{z_t = i} & \text{if}\quad c_t > 1\\
		\mathbf{A}(i, j) & \text{if}\quad c_t = 1
	\end{cases},
\end{align}
where $\delta_{\mathrm{cond}}$ denotes the delta function which takes the value 1 only when $\mathrm{cond}$ is true. 

Although SDSs with explicit switch duration distributions can identify long-term time-dependent switching patterns, the switch transitions are not informed by the state---inhibiting their ability to model state-dependent switching events. Furthermore, to the best of our knowledge, SDSs with explicit duration models have only been studied for Gaussian linear states~\cite{chiappa2010movement,chiappa2019explicit,oh2008learning}.

\section{Recurrent explicit duration switching dynamical systems}
\label{sec:model}

In this section we describe the Recurrent Explicit Duration Switching Dynamical System (RED-SDS) that combines both state-to-switch recurrence and explicit duration modeling for switches in a single non-linear model. 
We begin by formulating the generative model as a recurrent switching dynamical system that explicitly models the switch durations using increasing count variables. We then discuss how to perform efficient inference for different sets of latent variables. Finally, we discuss how to estimate the parameters of RED-SDS using maximum likelihood.

\subsection{Model formulation}
\label{subsec:model-formulation}
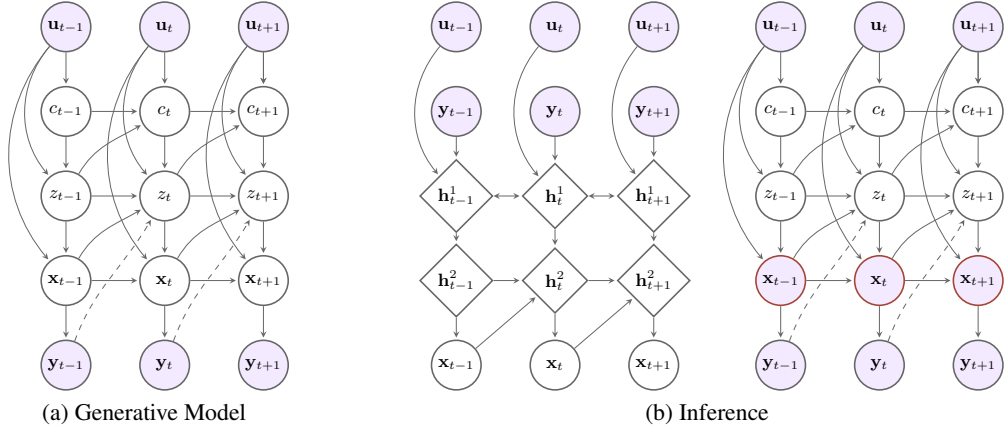
\begin{figure}[htb]
	\centering
	\subfloat[Generative Model]{\scalebox{0.75}{\begin{tikzpicture}[shorten >=1pt,->,draw=black!60, font=\footnotesize, scale=0.5]%
			\definecolor{lilac}{HTML}{e4cbff}
			\tikzstyle{every edge}=[draw,-stealth, thin];
			\tikzstyle{control node}=[circle, draw, thick, fill=lilac!40, minimum width=25pt, inner sep=2pt];
			\tikzstyle{count node}=[circle, draw, thick, minimum size=25pt, inner sep=2pt];
			\tikzstyle{switch node}=[circle, draw, thick, minimum size=25pt, inner sep=2pt];
			\tikzstyle{state node}=[circle, draw, thick, minimum size=25pt, inner sep=2pt];
			\tikzstyle{observation node}=[circle, draw, thick, fill=lilac!40, minimum width=25pt, inner sep=2pt];
			
			\foreach \i/\j in {1/{t-1}, 2/{t}, 3/{t+1}} {%
				\node[control node] (u\i) at (3.5*\i,12) {$\mathbf{u}_{\j}$};%
				\node[count node] (c\i) at (3.5*\i,9) {$c_{\j}$};%
				\node[switch node] (z\i) at (3.5*\i,6) {$z_{\j}$};%
				\node[state node] (x\i) at (3.5*\i,3) {$\mathbf{x}_{\j}$};%
				\node[observation node] (y\i) at (3.5*\i,0) {$\mathbf{y}_{\j}$};%
				\path (z\i) edge (x\i);%
				\path (x\i) edge (y\i);%
			}%
			
			\foreach \i/\j in {1/2, 2/3} {%
				\path (u\i) edge (c\i);%
				\path (u\i) edge [out=230, in=130] (z\i);%
				\path (u\i) edge [out=230, in=130] (x\i);%
				\path (c\i) edge (z\i);%
				\path (z\i) edge [out=60, in=210] (c\j);%
				\path (c\i) edge (c\j) ;%
				\path (x\i) edge (x\j);%
				\path (z\i) edge (z\j);%
				\path (x\i) edge  [out=60, in=210] (z\j);%
				\path (y\i) edge  [out=70, in=240, dashed] (z\j);%
			}%
			\path (u3) edge (c3);%
			\path (c3) edge (z3);%
			\path (u3) edge (c3);%
			\path (u3) edge [out=230, in=130] (z3);%
			\path (u3) edge [out=230, in=130] (x3);%
			
		\end{tikzpicture}}}
	\hspace{4em}
	\subfloat[Inference]{\scalebox{0.75}{\begin{tikzpicture}[shorten >=1pt,->,draw=black!60, font=\footnotesize, scale=0.5]%
			\definecolor{lilac}{HTML}{e4cbff}
			\definecolor{brickred}{HTML}{8f1402}
			\tikzstyle{every edge}=[draw,-stealth, thin];
			\tikzstyle{control node}=[circle, draw, thick, fill=lilac!40, minimum width=25pt, inner sep=2pt];
			\tikzstyle{count node}=[circle, draw, thick, minimum size=25pt, inner sep=2pt];
			\tikzstyle{deterministic node}=[diamond, draw, thick, minimum size=33pt, inner sep=2pt];
			\tikzstyle{switch node}=[circle, draw, thick, minimum size=25pt, inner sep=2pt];
			\tikzstyle{state node}=[circle, draw, thick, minimum size=25pt, inner sep=2pt];
			\tikzstyle{infstate node}=[circle, draw=brickred!80, thick, fill=lilac!40, minimum size=25pt, inner sep=2pt];
			\tikzstyle{observation node}=[circle, draw, thick, fill=lilac!40, minimum width=25pt, inner sep=2pt];
			
				\foreach \i/\j in {1/{t-1}, 2/{t}, 3/{t+1}} {%
				\node[control node] (infu\i) at (3.5*\i,12) {$\mathbf{u}_{\j}$};%
				\node[observation node] (infy\i) at (3.5*\i,9) {$\mathbf{y}_{\j}$};%
				\node[deterministic node] (infh1\i) at (3.5*\i,6) {$\mathbf{h}^1_{\j}$};%
				\node[deterministic node] (infh2\i) at (3.5*\i,3) {$\mathbf{h}^2_{\j}$};%
				\node[state node] (infx\i) at (3.5*\i,0) {$\mathbf{x}_{\j}$};%
				\path (infy\i) edge (infh1\i);
				\path (infh1\i) edge (infh2\i);
				\path (infh2\i) edge (infx\i);
				\path (infu\i) edge  [out=230, in=130] (infh1\i);%
			}%
			
			\foreach \i/\j in {1/2, 2/3} {%
				\path (infh1\i) edge [stealth-stealth] (infh1\j);
				\path (infh2\i) edge (infh2\j);
				\path (infx\i) edge (infh2\j);
			}%
			
			\foreach \i/\j in {4/{t-1}, 5/{t}, 6/{t+1}} {%
				\node[control node] (u\i) at (1 + 3.5*\i,12) {$\mathbf{u}_{\j}$};%
				\node[count node] (c\i) at (1 + 3.5*\i,9) {$c_{\j}$};%
				\node[switch node] (z\i) at (1 + 3.5*\i,6) {$z_{\j}$};%
				\node[infstate node] (x\i) at (1 + 3.5*\i,3) {$\mathbf{x}_{\j}$};%
				\node[observation node] (y\i) at (1 + 3.5*\i,0) {$\mathbf{y}_{\j}$};%
				\path (z\i) edge (x\i);%
				\path (x\i) edge (y\i);%
			}%
			
			\foreach \i/\j in {4/5, 5/6} {%
				\path (u\i) edge (c\i);%
				\path (u\i) edge [out=230, in=130] (z\i);%
				\path (u\i) edge [out=230, in=130] (x\i);%
				\path (c\i) edge (z\i);%
				\path (z\i) edge [out=60, in=210] (c\j);%
				\path (c\i) edge (c\j) ;%
				\path (x\i) edge (x\j);%
				\path (z\i) edge (z\j);%
				\path (x\i) edge  [out=60, in=210] (z\j);%
				\path (y\i) edge  [out=70, in=240, dashed] (z\j);%
			}%
			\path (u6) edge (c6);%
			\path (c6) edge (z6);%
			\path (u6) edge (c6);%
			\path (u6) edge [out=230, in=130] (z6);%
			\path (u6) edge [out=230, in=130] (x6);%
			
\end{tikzpicture}}}
	\caption{\textbf{(a)} Forward generative model of RED-SDS. \textbf{(b)} Left: Approximate inference for the states $\vx_t$ using an inference network.
	$\mathbf{h}^1_t$ is given by a non-causal network and $\mathbf{h}^2_t$ is given by a causal RNN.
	Right: Exact inference for switch $z_t$ and count $c_t$ variables given pseudo-observations (highlighted in red) of $\vx_t$ provided by the inference network. (Shaded) circles represent (observed) random variables, diamonds represent deterministic nodes, and dashed lines represent optional connections.}
	\label{fig:REDSDS-inference}
\end{figure}

Consider the graphical model in Fig. \ref{fig:REDSDS-inference} (a); the joint distribution of the counts $c_t \in \{1, \dots, \dmax\}$, the switches $z_t \in \{1, \dots, K\}$, the states $\vx_t \in \mathbb{R}^m$, and the observations $\vy_t \in \mathbb{R}^d$, conditioned on the control inputs $\mathbf{u}_t \in \mathbb{R}^c$, factorizes as 
\begin{equation}
\begin{aligned}
	p_\theta&(\vy_{1:T}, \vx_{1:T}, z_{1:T}, c_{1:T} | \vu_{1:T}) 
	= p(\vy_1|\vx_1)p(\vx_1|z_1, \vu_1)p(z_1 | \vu_1) \\
	&\cdot\left[\prod_{t=2}^T 
	p(\vy_t|\vx_t)
	p(\vx_t|\vx_{t-1}, z_t, \vu_t)
	p(z_t| \vx_{t-1}, z_{t-1}, c_t, \vu_t)
	p(c_t|z_{t-1}, c_{t-1}, \vu_t)
	\right].
\end{aligned}
\label{eq:REDSDS-joint}
\end{equation}
Similar to~\cite{oh2008learning,chiappa2019explicit}, we consider increasing count variables $c_t$ to incorporate explicit switch durations into the model, i.e., $c_t$ can either increment by 1 or reset to 1 at every timestep and represent the run-length of the current regime.
A self-transition is allowed after the exhaustion of $\dmax$ steps for flexibility.
In the subsequent discussion we omit the control inputs $\vu_t$ for clarity of exposition.

We model the initial prior distributions in Eq. \eqref{eq:REDSDS-joint}  for the respective discrete and continuous case as 
\begin{align}
	p(z_1) &= \mathrm{Cat}(z_1; \bm{\pi}),\\
	p(\vx_1 | z_1) &= \mathcal{N}(\vx_1; \bm{\mu}_{z_1}, \bm{\Sigma}_{z_1}),
\end{align}
where $\mathrm{Cat}$ denotes a categorical and $\mathcal{N}$ a multivariate Gaussian distribution.
The transition distributions for the discrete variables (count and switch) are given by 
\begin{align}
	p(c_t| z_{t-1}, c_{t-1}) &= 
	\begin{cases}
		v_{z_{t-1}}(c_{t-1}) & \text{if}\quad c_t = c_{t-1} + 1\\
		1 - v_{z_{t-1}}(c_{t-1}) & \text{if}\quad c_t = 1
	\end{cases},\\
	p(z_t| \vx_{t-1}, z_{t-1}, c_t ) &= 
	\begin{cases}
		\delta_{z_t = z_{t-1}} & \text{if}\quad c_t > 1\\
		\mathrm{Cat}(z_t; \mathcal{S}_{\tau}(f_z(\vx_{t-1}, z_{t-1}))) & \text{if}\quad c_t = 1
	\end{cases},
\end{align}
where $\mathcal{S}_{\tau}$ is the tempered softmax function (cf. Section \ref{subsec:learning}) with temperature $\tau$, and $f_z$ can be a linear function or a neural network. 
The probability of a count increment $v_k$ for a switch $k$ is defined via the duration model $\rho_k$ as in Eq. \eqref{eq:count_incr}.
The continuous state transition and the emission are given by 
\begin{align}
	p(\vx_t|\vx_{t-1}, z_{t}) &= \mathcal{N}(\vx_t; f^\mu_x(\vx_{t-1}, z_{t}), f^\Sigma_x(\vx_{t-1}, z_{t})),\\
	p(\vy_t|\vx_t) &= \mathcal{N}(\vy_t; f_y^\mu(\vx_t), f_y^\Sigma(\vx_t)),
\end{align} 
where $f_x^\mu$, $f_x^\Sigma$, $f_y^\mu$, $f_y^\Sigma$ are again linear functions or neural networks. 

The model is general and flexible enough to handle both state- and time-dependent switching. 
The switch transitions $z_{t-1} \to z_t$ are conditioned on the previous state $\vx_{t-1}$ which ensures that the switching events occur in a ``closed loop''. %
The switch duration models $\rho_k$ provide flexibility to stay long term in the same regime, allowing to better capture time-dependent switching. 
We use increasing count variables to incorporate switch durations into our model as they are more amenable to the case when the count transitions depend on the control $\vu_t$. 
For instance, decreasing count variables, another popular option~\cite{dai2016recurrent,liu2018structured,chiappa2019explicit}, \emph{deterministically} count down from the sampled segment duration length to 1. 
This makes it difficult to condition the switch duration model on the control inputs. In contrast, increasing count variables increment or reset probabilistically at every timestep.

\subsection{Inference}
\label{subsec:inference}
Exact inference is intractable in SDSs and scales exponentially with time~\cite{lerner2002hybrid}. Various approximate inference procedures have been developed for traditional SDSs~\cite{doucet2000rao,ghahramani2000variational,barber2006ecslds}, while more recently inference networks have been used for amortized inference for all or a subset of latent variables~\cite{Johnson2016ComposingGM,kim2019variational,dong2020collapsed,kurle2020deeprao}. Particularly, \citet{dong2020collapsed} used an inference network for the states and performed exact HMM-like inference for the switches, conditioned on the states. We take a similar approach and use an inference network for the continuous latent variables (states) and perform conditionally exact inference for the discrete latent variables (switches and counts) similar to the forward-backward procedure for HSMM~\cite{yu2010hidden,chiappa2019explicit}.
We define the variational approximation to the true posterior $p(\vx_{1:T},  z_{1:T}, c_{1:T} |\vy_{1:T})$ as $q(\vx_{1:T}, z_{1:T}, c_{1:T} |\vy_{1:T}) = q_\phi(\vx_{1:T}|\vy_{1:T})p_\theta(z_{1:T}, c_{1:T} | \vy_{1:T}, \vx_{1:T})$ where $\phi$ and $\theta$ denote the parameters of the inference network and the generative model respectively.

\textbf{Approximate inference for states.} The posterior distribution of the states, $q_\phi(\vx_{1:T}|\vy_{1:T})$, is approximated using an inference network. We first process the observation sequence $\vy_{1:T}$ using a non-causal network such as a bi-RNN or a Transformer~\cite{Vaswani2017AttentionIA} to simulate smoothing by incorporating both past and future information. The non-causal network returns an embedding of the data $\mathbf{h}^1_{1:T}$ which is then fed to a causal RNN that outputs the posterior distribution $q_\phi(\vx_{1:T}|\vy_{1:T}) = \prod_t q(\vx_{t}|\vx_{1:t-1}, \mathbf{h}^1_{1:T})$. See Fig. \ref{fig:REDSDS-inference} (b) for an illustration of the inference procedure. %

\textbf{Exact inference for counts and switches.} Inference for the switches $z_{1:T}$ and the counts $c_{1:T}$ can be performed exactly conditioned on states $\vx_{1:T}$ and observations $\vy_{1:T}$. Samples from the approximate posterior $\tilde{\vx}_{1:T} \sim q(\vx_{1:T}|\vy_{1:T})$ are used as pseudo-observations of $\vx_{1:T}$ to infer the posterior distribution $p_\theta(z_{1:T}, c_{1:T} | \vy_{1:T}, \tilde{\vx}_{1:T})$. A naive approach to infer this distribution is by treating the pair $(c_t, z_t)$ as a ``meta switch'' that takes $K\dmax$ possibles values and perform HMM-like forward-backward inference. However, this results in a computationally expensive $O(TK^2\dmax^2)$ procedure that scales poorly with $\dmax$. Fortunately, we can pre-compute some terms in the forward-backward equations by exploiting the fact that the count variable can only increment by 1 or reset to 1 at every timestep. This results in an $O(TK(K+\dmax))$ algorithm that scales gracefully with $\dmax$~\cite{chiappa2019explicit}. The forward $\alpha_t$ and backward $\beta_t$ variables, defined as
\begin{align}
	\alpha_t(z_t, c_t) &= p( \vy_{1:t}, \vx_{1:t}, z_t, c_t),\\
	\beta_t(z_t, c_t) &= p(\vy_{t+1:T}, \vx_{t+1:T} | \vx_t, z_t, c_t),
\end{align} 
can be computed by modifying the forward-backward recursions used for the HSMM~\cite{chiappa2019explicit} to handle the additional observed variables $\vx_{1:t}$. We refer the reader to Appendix \ref{app:fwd_bwd} for the exact derivation. %

\subsection{Learning}
\label{subsec:learning}

The parameters $\{\phi, \theta\}$ can be learned by maximizing the evidence lower bound (ELBO):
\begin{equation}
\begin{aligned}
	\mathcal{L}_\mathrm{ELBO} &= \mathbb{E}_{q(\vx_{1:T}|\vy_{1:T})p(z_{1:T}, c_{1:T} | \vy_{1:T}, \vx_{1:T})}\left[\log\frac{p(\vy_{1:T}, \vx_{1:T}, z_{1:T}, c_{1:T}, )}{q(\vx_{1:T}|\vy_{1:T})p(z_{1:T}, c_{1:T} | \vy_{1:T}, \vx_{1:T})}\right]\\
	&= \mathbb{E}_{q(\vx_{1:T}|\vy_{1:T})}\left[\log\frac{p(\vy_{1:T}, \vx_{1:T})}{q(\vx_{1:T}|\vy_{1:T})}\right].
\end{aligned}
\end{equation}

The likelihood term $p(\vy_{1:T}, \vx_{1:T})$ can be computed using the forward variable $\alpha_T(z_T, c_T)$ by marginalizing out the switches and the counts,
\begin{align}
	p(\vy_{1:T}, \vx_{1:T}) &= \sum_{z_T, c_T}\alpha_T(z_T, c_T), \label{eq:marginalize-discrete}
\end{align}
and the entropy term $-\mathbb{E}_{q(\vx_{1:T}|\vy_{1:T})}\left[\log q(\vx_{1:T}|\vy_{1:T})\right]$ can be computed using the approximate posterior $q(\vx_{1:T}|\vy_{1:T})$ output by the inference network. The ELBO can be maximized via stochastic gradient ascent given that the posterior $q(\vx_{1:T}|\vy_{1:T})$ is reparameterizable.

We note that ~\citet{dong2020collapsed} used a lower bound for the likelihood term in Switching Non-Linear Dynamical Systems (SNLDS);
however, it can be computed succinctly by marginalizing out the discrete random variable (i.e., the switch in SNLDS) from the forward variable $\alpha_T$, similar to Eq. \eqref{eq:marginalize-discrete}. 
Using our objective function, we observed that the model was less prone to posterior collapse (where the model ends up using only one switch) and we did not require the additional ad-hoc KL regularizer used in \citet{dong2020collapsed}. Please refer to Appendix \ref{app:seg-baselines} for a brief discussion on the likelihood term in SNLDS.

\textbf{Temperature annealing.} We use the tempered softmax function $\mathcal{S}_\tau$ to map the logits to probabilities for the switch transition $p(z_t | \vx_{t-1},  z_{t-1}, c_t=1)$ and the duration models $\rho_k(d)$ which is defined as
\begin{align}
	\mathcal{S}_\tau(\mathbf{o})_i &= \frac{\exp\left(\frac{o_i}{\tau}\right)}{\sum_j \exp\left(\frac{o_j}{\tau}\right)},
	\label{eq:temp_softmax}
\end{align}
where $\mathbf{o}$ is a vector of logits. The temperature $\tau$ is deterministically annealed from a high value during training. The initial high temperature values soften the categorical distribution and encourage the model to explore all switches and durations. This prevents the model from getting stuck in poor local minima that ignore certain switches or longer durations which might explain the data better.

\section{Related work}
\label{sec:related}

The most relevant components of RED-SDS are recurrent state-to-switch connections and the explicit duration model, enabling both for state- and time-dependent switching. 
Additionally, RED-SDS allows for efficient approximate inference (analytic for switches and counts), despite parameterizing the various conditional distributions through neural networks. Existing methods address only a subset of these features as we discuss in the following.

The most prominent SDS is the Switching Linear Dynamical System (SLDS), where each regime is described by linear dynamics and additive Gaussian noise. 
A major focus of previous work has been on efficient approximate inference algorithms that exploit the Gaussian linear substructure (e.g.,~\cite{ghahramani2000variational,zoeter2005changepoint,doucet2000rao}). In contrast to RED-SDS, these models lack recurrent state-to-switch connections and duration variables and are limited to linear regimes.

Previous work has addressed the state-dependent switching by introducing a connection to the continuous state of the dynamical system \cite{barber2006ecslds, linderman2016recurrent, becker2019sdvbf, kurle2020deeprao}. 
The additional recurrence complicates inference w.r.t.\ the continuous states;
prior work uses expensive sampling methods in order to approximate the corresponding integrals~\cite{barber2006ecslds} or as part of a message passing algorithm for joint inference of states and parameters~\cite{linderman2016recurrent}. 
On the other hand, ARSGLS~\cite{kurle2020deeprao} avoids sampling the continuous states by using conditionally linear state-to-switch connections and softmax-transformed Gaussian switch variables.
However, both the ARSGLS and the related KVAE \cite{fraccaro2017disentangled} can be interpreted as an SLDS with ``soft'' switches that interpolate linear regimes continuously rather than truly discrete states. 
This makes them less suited for time series segmentation compared to RED-SDS. 
Contrary to the aforementioned models, RED-SDS 
allows non-linear regimes described by neural networks and 
incorporates a discrete explicit duration model without complicating inference w.r.t.\ the continuous states, since closed-form expressions are used for the discrete variables instead. Using amortized variational inference for continuous variables and analytic expressions for discrete variables has been proposed previously for segmentation in SNLDS~\cite{dong2020collapsed}. 
RED-SDS extends this via an additional explicit duration variable that represents the run-length for the currently active regime. 

Explicit duration variables have previously been proposed for changepoint detection~\cite{adams2007bayesian,agudelo2019bayesian} and segmentation~\cite{chiappa2010movement,johnson2012hierarchical}. 
For instance, BOCPD~\cite{adams2007bayesian} is a Bayesian online changepoint detection model with explicit duration modeling. 
RED-SDS improves upon BOCPD by allowing for segment labeling rather than just detecting changepoints. The HDP-HSMM~\cite{johnson2012hierarchical} is a Bayesian non-parametric extension to the traditional HSMM. Recent work~\cite{dai2016recurrent,liu2018structured} has also combined HSMM with RNNs for amortized inference. These models---being variants of HSMM---do not model the latent dynamics of the data like RED-SDS. \citet{chiappa2010movement} proposed approximate inference techniques for a variant of SLDS with explicit duration modeling. In contrast, RED-SDS is a more general non-linear model that allows for efficient amortized inference---closed-form w.r.t. the discrete latent variables.

\section{Experiments}
\label{sec:exp}

In this section, we present empirical results on two prominent 
time series tasks: segmentation and forecasting. Our primary goals were to determine if RED-SDS (a) can discover meaningful switching patterns in the data in an unsupervised manner, and (b) can probabilistically extrapolate a sequence of observations, serving as a viable generative model for forecasting. In the following, we discuss the main results and relegate details to the appendix.
\subsection{Segmentation}

\begin{wrapfigure}[27]{r}{0.6\textwidth}
    \vspace{-7mm}
	\subfloat[{\small Bouncing ball}]{\includegraphics[width=\linewidth]{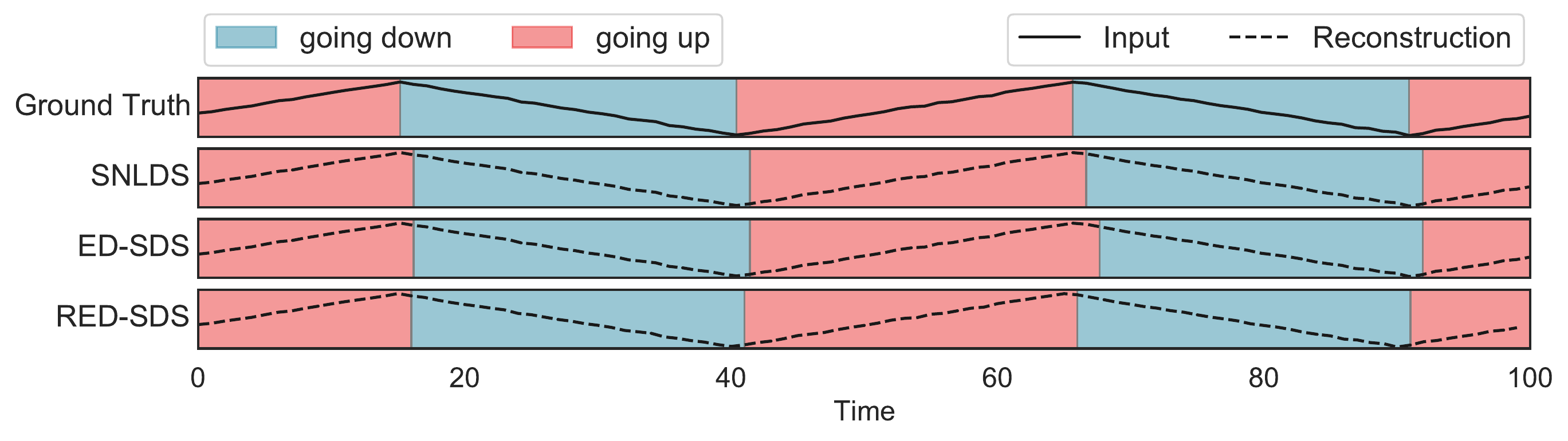}}\hfill \vspace{-2mm}
	\subfloat[3 mode system]{\includegraphics[width=\linewidth]{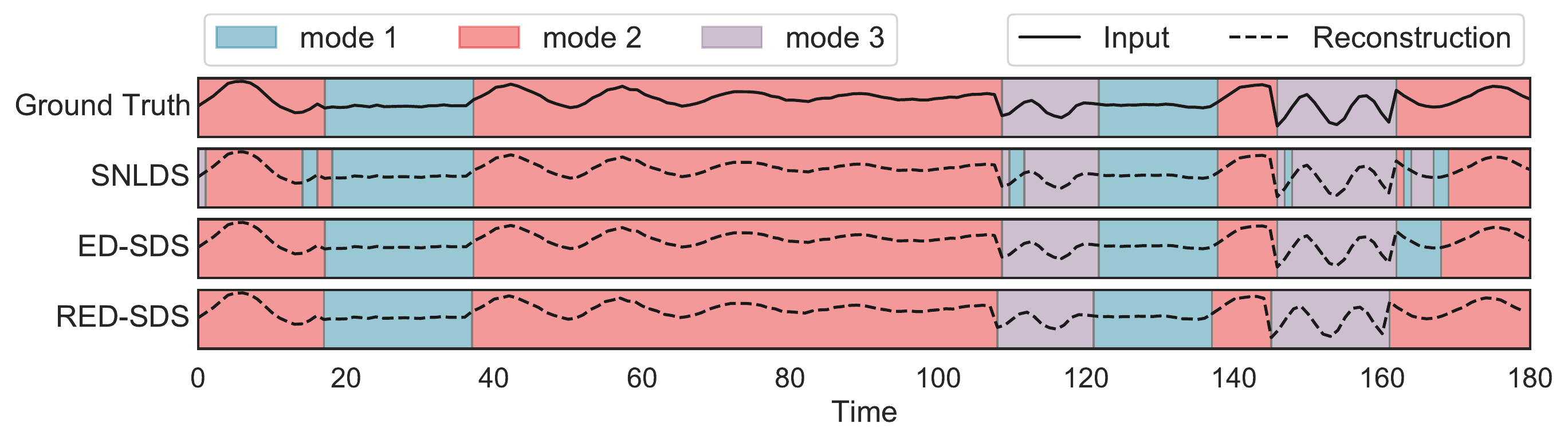}}\hfill \vspace{-2mm}
	\subfloat[Dancing bees]{\includegraphics[width=\linewidth]{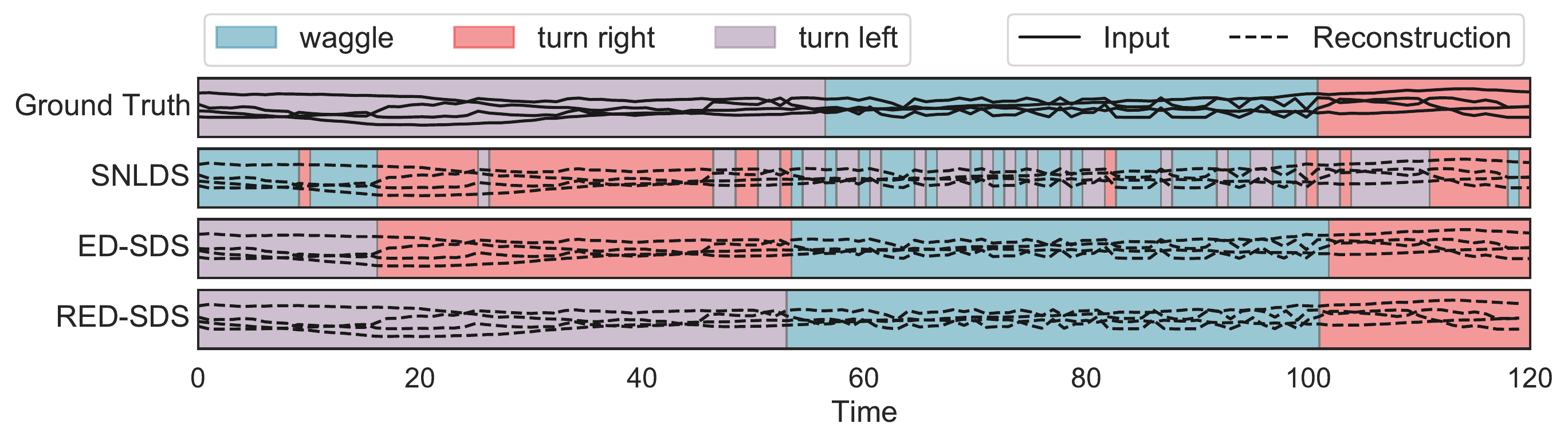}}\hfill \vspace{-1mm}
	\caption{\small Qualitative segmentation results on the bouncing ball, 3 mode system, and dancing bees datasets. 
	Background colors represent the different operating modes.
	}
	\label{fig:seg}
\end{wrapfigure}

We experimented with two instantiations of our model: \REDSDS{} (complete model) and \EDSDS{}, the ablated variant without state-to-switch recurrence.
We compared against the closely related \SNLDS{}~\cite{dong2020collapsed} trained with a modified objective function.
The original objective proposed in~\citep{dong2020collapsed} suffered from training difficulties: it resulted in frequent posterior collapse and was sensitive to the cross-entropy regularization term.
Our version of \SNLDS{} can be seen as a special case of \REDSDS{} with $\dmax = 1$, i.e., without the explicit duration modeling (cf. Appendix \ref{app:seg-baselines}).
We also conducted preliminary experiments on soft-switching models: \KVAE{}~\cite{fraccaro2017disentangled} and \ARSGLS{}~\cite{kurle2020deeprao}.
However, these models use a continuous interpolation of the different operating modes which cannot always be correctly assigned to a single discrete mode, hence we do not report these unfavorable findings here (cf. Appendix \ref{app:seg-baselines}).
For all models, we performed segmentation by taking the most likely value of the switch at each timestep from the posterior distribution over the switches.
As the segmentation labels are arbitrary and may not match the ground truth labels, we evaluated the models using multiple metrics: frame-wise segmentation accuracy (after matching the labelings using the Hungarian algorithm~\cite{kuhn1955hungarian}), Normalized Mutual Information (NMI)~\cite{vinh2010information}, and Adjusted Rand Index (ARI)~\cite{hubert1985comparing} (cf. Appendix~\ref{app:seg-metrics}).

We conducted experiments on three benchmark datasets: bouncing ball, 3 mode system, and dancing bees to investigate different segmentation capabilities of the models. We refer the reader to Appendix \ref{app:seg_datasets} for details on how these datasets were generated/preprocessed. For all the datasets, we set the number of switches equal to the number of ground truth operating modes.

\textbf{Bouncing ball.} We generated the bouncing ball dataset similar to~\cite{dong2020collapsed}, which comprises univariate time series that encode the location of a ball bouncing between two fixed walls with a constant velocity and elastic collisions.
The underlying system switches between two operating modes (going up/down) and the switching events are completely governed by the state of the ball, i.e., a switch occurs only when the ball hits a wall.
As such, the switching events are best explained by state-to-switch recurrence.
All models are able to segment this simple dataset well as shown qualitatively in Fig \ref{fig:seg} (a) and quantitatively in Table \ref{tab:results_segmentation}. We note that despite the seemingly qualitative equivalence, models with state-to-switch recurrence perform best quantitatively.
\REDSDS{} learns to ignore the duration variable by assigning almost all probability mass to shorter durations (cf. Appendix \ref{app:seg-add-results}), which is intuitive since the recurrence best explains this dataset.

\begin{wrapfigure}[19]{r}{0.6\textwidth}
	\vspace{-8.5mm}
	\subfloat[True duration model]{\includegraphics[width=\linewidth]{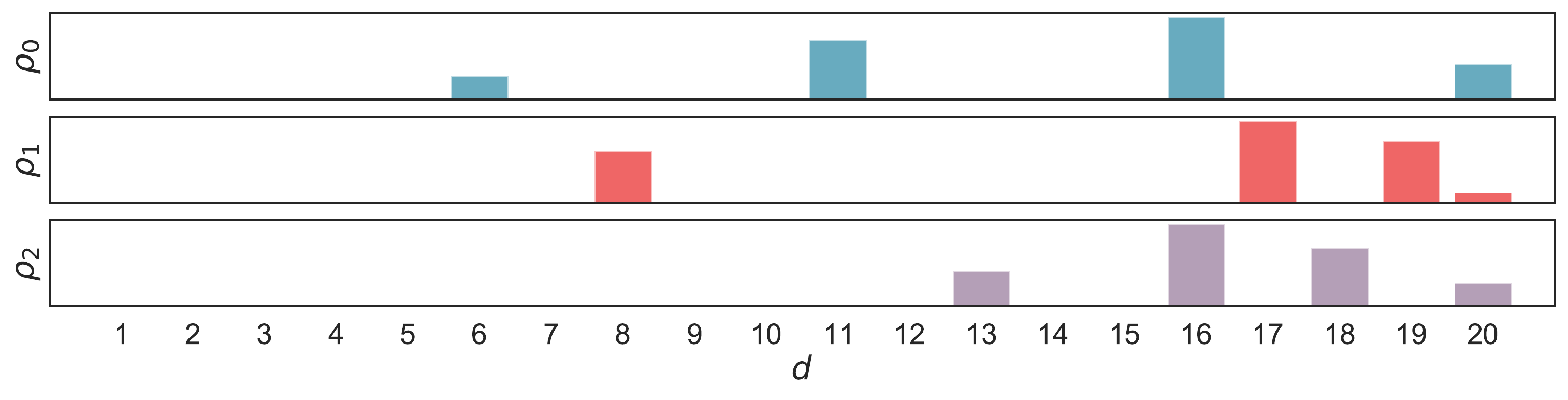}}\hfill \vspace{-2.5mm}
	\subfloat[Learned duration model]{\includegraphics[width=\linewidth]{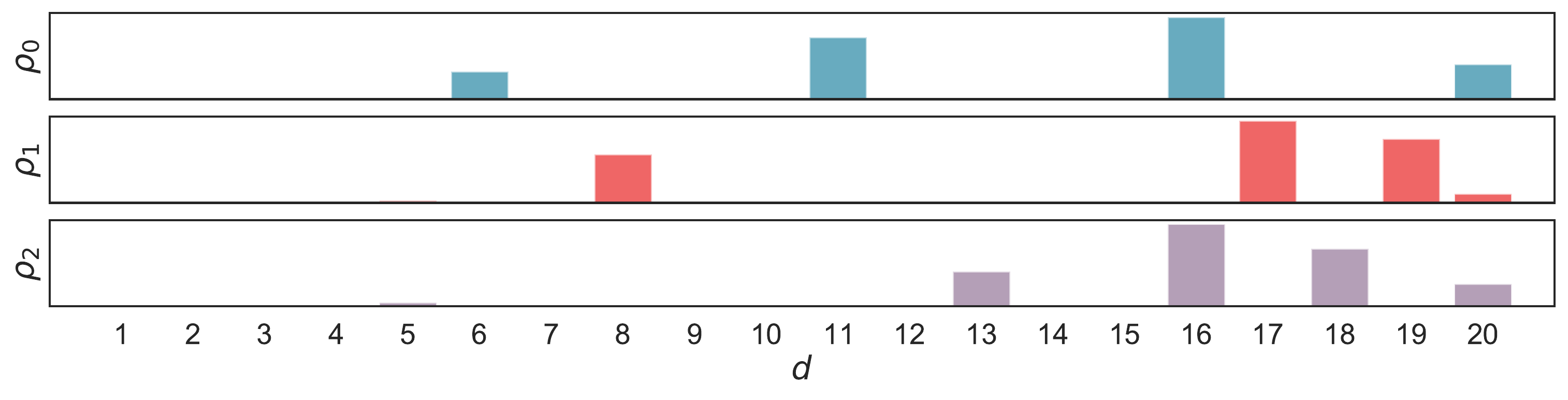}}\hfill \vspace{-1mm}
	\caption{\small The ground truth duration model for the 3 mode system dataset (top) and the duration model learned by \REDSDS{} (bottom). The x-axis represents the durations from 1 to 20 and the y-axis represents the duration probabilities of the 3 modes $\rho_0(d), \rho_1(d)$, and $\rho_2(d)$.}
	\label{fig:count-slds-seg}
	\vspace{-8mm}
\end{wrapfigure}

\textbf{3 mode system.} We generated this dataset from a switching linear dynamical system with 3 operating modes and an explicit duration model for each mode (shown in Fig. \ref{fig:count-slds-seg} (a)).
We study this dataset in the context of time-dependent switching---the operating mode switches after a specific amount of time elapses based on its duration model.
Both \EDSDS{} and \REDSDS{} learn to segment this dataset almost perfectly as shown in Fig. \ref{fig:seg} (b) and Table \ref{tab:results_segmentation} owing to their ability to explicitly model switch durations.
In contrast, \SNLDS{} fails to completely capture the long-term temporal patterns, resulting in spurious short-term segments as shown in Fig. \ref{fig:seg} (b).
Moreover, \REDSDS{} is able to recover the duration models associated with the different modes (Fig. \ref{fig:count-slds-seg}).
These results demonstrate that explicit duration models can better identify the time-dependent switching patterns in the data and can leverage prior knowledge about the switch durations imparted via the $\dmin$ and $\dmax$ hyperparameters.

\begin{table*}
	\footnotesize
	\centering
	\caption{\small Quantitative results on segmentation tasks. Accuracy, NMI, and ARI denote the frame-wise segmentation accuracy, the Normalized Mutual Information, and the Adjusted Rand Index metrics respectively (higher values are better). Mean and standard deviation are computed over 3 independent runs.}
	\resizebox{\columnwidth}{!}{\begin{tabular}{llcccc}
		\toprule
		& &      \bb &        \edslds  &      \bee & \bee\texttt{(K=2)} \\
		\midrule
		\multirow{3}{*}{Accuracy} & \SNLDS & \textbf{0.97\textpm0.00} & 0.82\textpm0.08 & 0.44\textpm0.01 & 0.63\textpm0.02 \\
		& \EDSDS~(ours) & 0.95\textpm0.00 & 0.97\textpm0.00 & 0.56\textpm0.06 & 0.79\textpm0.09 \\
		 & \REDSDS~(ours) & \textbf{0.97\textpm0.00} &\textbf{ 0.98\textpm0.00} & \textbf{0.73\textpm0.10} & \textbf{0.91\textpm0.04}  \\
		\midrule
		\multirow{3}{*}{NMI} & \SNLDS & \textbf{0.83\textpm0.01} &  0.63\textpm0.08  & 0.10\textpm0.04 & 0.05\textpm0.02 \\
		& \EDSDS~(ours) & 0.71\textpm0.00 & 0.89\textpm0.01 & 0.28\textpm0.02 & 0.31\textpm0.17 \\
		 & \REDSDS~(ours) &  0.81\textpm0.00 & \textbf{ 0.91\textpm0.01} &  \textbf{0.48\textpm0.07} & \textbf{0.60\textpm0.09} \\
		\midrule
		\multirow{3}{*}{ARI} & \SNLDS &  \textbf{0.90\textpm0.01} & 0.67\textpm0.11 & 0.10\textpm0.03 & 0.07\textpm0.02 \\
		& \EDSDS~(ours) & 0.81\textpm0.01 & 0.93\textpm0.00 & 0.27\textpm0.04 & 0.36\textpm0.19 \\
		 & \REDSDS~(ours) & 0.88\textpm0.00 & \textbf{0.95\textpm0.01} & \textbf{0.53\textpm0.11} & \textbf{0.68\textpm0.11} \\
		\bottomrule
	\end{tabular}}
	\vspace{-1.5em}
	\label{tab:results_segmentation}
\end{table*}

\textbf{Dancing bees.} We used the publicly-available dancing bees dataset~\cite{oh2008learning}---a challenging dataset that exhibits long-term temporal patterns and has been studied previously in the context of time series segmentation~\cite{Oh2005LearningAI,oh2006parameterized,Fox2008NonparametricBL}.
The dataset comprises trajectories of six dancer honey bees performing the waggle dance.
Each trajectory consists of the 2D coordinates and the heading angle of a bee at every timestep with three possible types of motion: waggle, turn right, and turn left.
Fig. \ref{fig:seg} (c) shows that \REDSDS{} is able to segment the complex long-term motion patterns quite well. In contrast, \EDSDS{} identifies the long segment durations but often infers the mode inaccurately while \SNLDS{} struggles to learn the long-term motion patterns resulting in oversegmentation.
This limitation of \SNLDS{} is particularly apparent in the ``waggle'' phase of the dance which involves rapid, shaky motion.
We also observed that sometimes \EDSDS{} and \REDSDS{} combined the turn right and turn left motions into a single switch, effectively segmenting the time series into regular (turn right and turn left) and waggle motion.
This results in another reasonable segmentation, particularly in the absence of ground-truth supervision.
We thus reevaluated the results after combining the turn right and turn left labels into a single label and present these results under \bee\texttt{(K=2)} in Table \ref{tab:results_segmentation}.
Empirically, \REDSDS{} significantly outperforms \EDSDS{} and \SNLDS{} on both labelings of the dataset. This suggests that real-world phenomena are better modeled by a combination of state- and time-dependent modeling capacities via state-to-switch recurrence and explicit durations, respectively.

\subsection{Forecasting}

\begin{table*}
	\footnotesize
	\center
	\caption{\small
		CRPS metrics (lower is better). 
		Mean and standard deviation are computed over 3 independent runs. 
		The method achieving the best result is highlighted in \textbf{bold}.}
	
	\resizebox{\columnwidth}{!}{\begin{tabular}{lccccc}
		\toprule
		&       \exchange &          \solar &           \electricity  &        \traffic &                 \wiki   \\
		\midrule
		\DeepAR &  0.019\textpm0.002 &  0.440\textpm0.004 &  \textbf{0.062\textpm0.004} &  0.138\textpm0.001 &  0.855\textpm0.552 \\
		\DeepState  & 0.017\textpm0.002 &  0.379\textpm0.002 &  0.088\textpm0.007 &  0.131\textpm0.005 &  0.338\textpm0.017 \\
		\KVAEMC  & 0.020\textpm0.001 & 0.389\textpm0.005  & 0.318\textpm0.011 & 0.261\textpm0.016 & 0.341\textpm0.032  \\
		\KVAERB  & 0.018\textpm0.001 &  0.393\textpm0.006  & 0.305\textpm0.022 & 0.221\textpm0.002 & 0.317\textpm0.013  \\
		\RSGLSISSM  &  0.014\textpm0.001 &   \textbf{0.358\textpm0.001} &  0.091\textpm0.004 &  0.206\textpm0.002 &  0.345\textpm0.010   \\
		\ARSGLS  & 0.022\textpm0.001 &  0.371\textpm0.007  & 0.154\textpm0.005 & 0.175\textpm0.008 & \textbf{0.283\textpm 0.006}  \\
		\REDSDS~(ours) & \textbf{0.013\textpm0.001} & 0.419\textpm0.010 & 0.066\textpm0.002 & \textbf{0.129\textpm0.002} &  0.318\textpm0.006\\
		\bottomrule
	\end{tabular}}
	\label{tab:results_crps}
\end{table*}

We evaluated \REDSDS\ in the context of time series forecasting on 5 popular public datasets available in GluonTS~\cite{alexandrov2020gluonts}, following the experimental set up of \cite{kurle2020deeprao}.
The datasets have either hourly or daily frequency with various seasonality patterns such as daily, weekly, or composite.
In Appendix \ref{app:for_datasets} we provide a detailed description of the datasets. 
We compared \REDSDS\ to closely related forecasting models: \ARSGLS\ and its variant \RSGLSISSM\ \cite{kurle2020deeprao}; \KVAEMC\ and \KVAERB, which refer to the original KVAE~\cite{fraccaro2017disentangled} and its Rao-Blackwellized variant (as described in \cite{kurle2020deeprao}) respectively; \DeepState\ \cite{rangapuram2018deep}; and \DeepAR{}~\cite{salinas2020deepar}, a strong \emph{discriminative} baseline that uses an autoregressive RNN (cf. Appendix \ref{app:forecast-baselines} for a discussion on these baselines). 

\begin{wrapfigure}[23]{r}{0.6\textwidth}
	\vspace{-4.5mm}
	\subfloat[$K=2$]{\includegraphics[width=\linewidth]{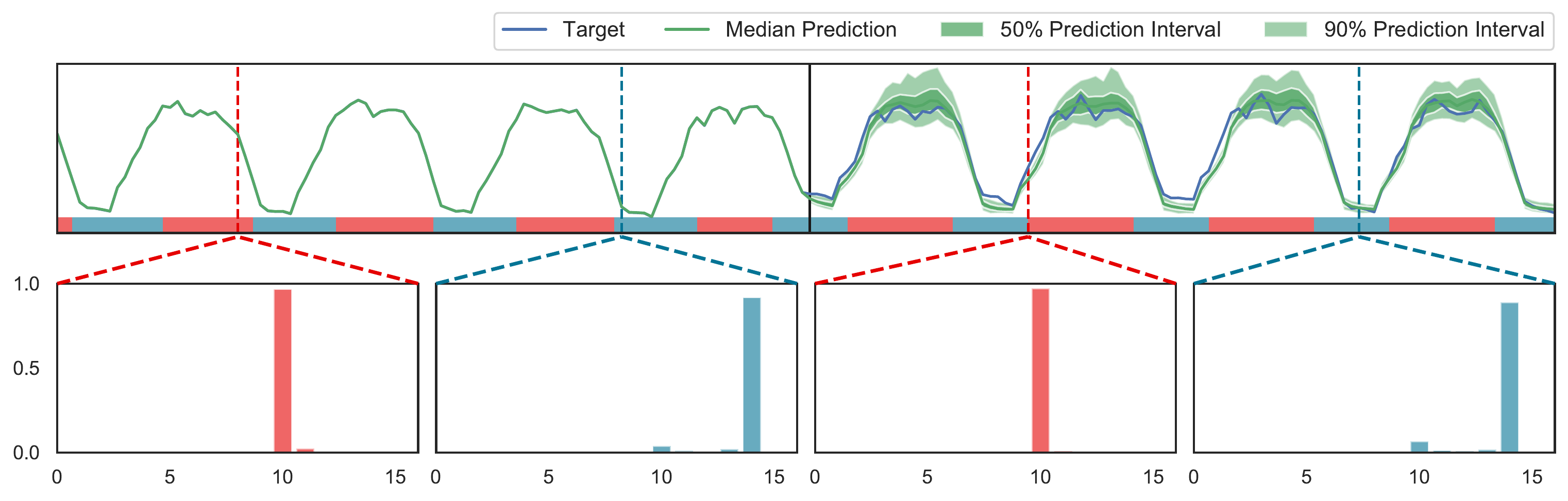}}\vspace{-1mm}
	\subfloat[$K=3$]{\includegraphics[width=\linewidth]{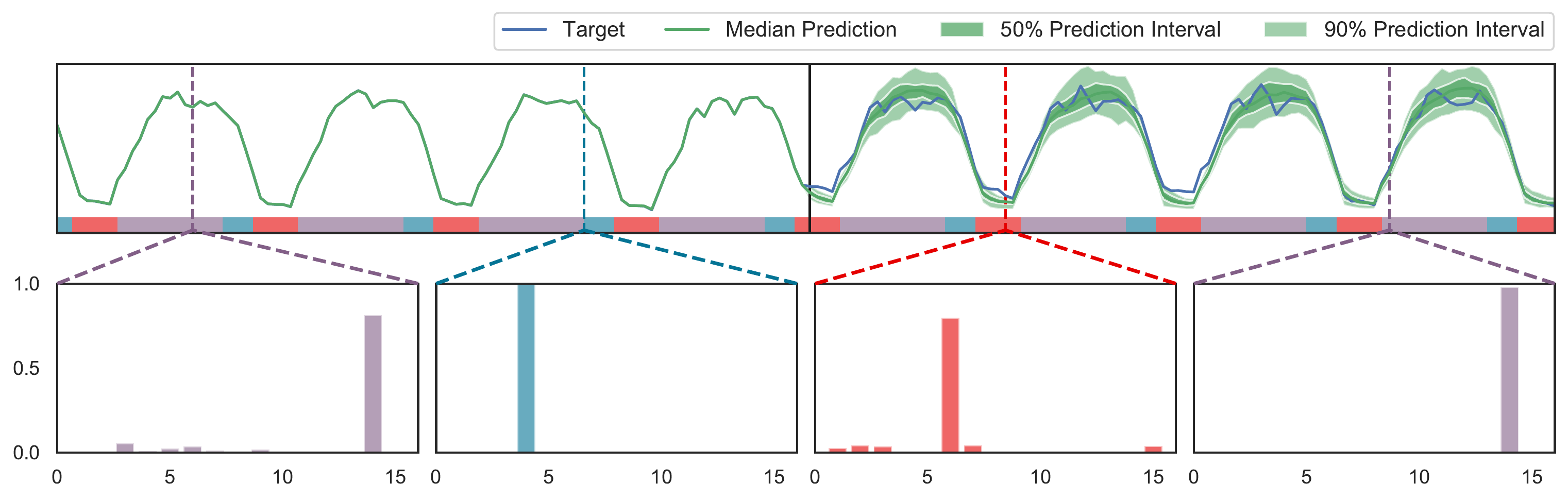}}\vspace{-0.1mm}
	\caption{\small  Segmentation and forecasting on an \electricity\ time series for (a) $K=2$ and (b) $K=3$ switches. The black vertical line indicates the start of forecasting. The plots at the second row of each figure indicate the duration model at the timestep marked by the corresponding vertical dashed lines.}
	\label{fig:forecast-plot}
\end{wrapfigure}

We used data prior to a fixed forecast date for training and test the forecasts on the remaining unseen data; the probabilistic forecasts are conditioned on the training range and computed with 100 samples for each method. 
We used a forecast window of 150 days and 168 hours for datasets with daily and hourly frequency, respectively.
We evaluated the forecasts using the \textit{continuous ranked probability score} (CRPS) \cite{matheson1976scoring}, a proper scoring rule \cite{gneiting2007strictly} 
(cf. Appendix  \ref{app:crps}).
The results are reported in Table \ref{tab:results_crps}; \REDSDS{} compares favorably or competitively to the baselines on 4 out of 5 datasets.

Figure \ref{fig:forecast-plot} illustrates how \REDSDS\ can infer meaningful switching patterns from the data and extrapolate the learned patterns into the future. 
It perfectly reconstructs the past of the time series and segments it in an interpretable manner without an imposed seasonality structure, e.g., as used in \DeepState\ and \RSGLSISSM.
The same switching pattern is consistently predicted into the future, simplifying the forecasting problem by breaking the time series into different regimes with corresponding properties such as trend or noise variance. Further, the duration models at several timesteps (the duration model is conditioned on the control $\vu_t$) indicate that the model has learned how long each regime lasts and therefore avoids oversegmentation which would harm the efficient modeling of each segment. Notably, the model learns meaningful regime durations that sum up to the 24-hour day/night period for both $K=2$ and $K=3$ switches. Thus, \REDSDS\ brings the added benefit of interpretability---both in terms of the discrete operating mode and the segment durations---while obtaining competitive quantitative performance relative to the baselines.

\section{Conclusion and future work}
\label{sec:conclusion}
Many real-world time series exhibit prolonged regimes of consistent dynamics as well as persistent statistical properties for the durations of these regimes. By explicitly modeling both state- and time-dependent switching dynamics, our proposed RED-SDS can more accurately model such data. Experiments on a variety of datasets show that RED-SDS---when equipped with an efficient inference algorithm that combines amortized variational inference with exact inference for continuous and discrete latent variables---improves upon existing models on segmentation tasks, while performing similarly to strong baselines for forecasting.

One current challenge of the proposed model is that learning interpretable segmentation sometimes requires careful hyperparameter tuning (e.g., $\dmin$ and $\dmax$). 
This is not surprising given the flexible nature of the neural networks used as components in the base dynamical system.
A promising future research direction is to incorporate simpler models that have a predefined structure, thus exploiting domain knowledge. 
For instance, many forecasting models such as DeepState and RSGLS-ISSM parametrize classical level-trend and seasonality models in a non-linear fashion. 
Similarly, simple forecasting models with such structure could be used as base dynamical systems along with more flexible neural networks.
Another interesting application is semi-supervised time series segmentation. 
For timesteps where the correct regime label is known, it is straightforward to condition on this additional information rather than performing inference; this may improve segmentation accuracy while providing an inductive bias that corresponds to an interpretable segmentation.

\section*{Funding disclosure}

This work was funded by Amazon Research. This research is supported in part by the National Research Foundation Singapore under its AI Singapore Programme (Award Number: AISG-RP-2019-011) to H. Soh.

\bibliographystyle{plainnat}
\bibliography{refs.bib}
\newpage

\appendix

\section{Model: additional details}

\subsection{Forward-backward algorithm}
\label{app:fwd_bwd}

As mentioned in Section \ref{subsec:inference}, inference for the discrete latent variables, i.e., the counts $c_{1:T}$ and the switches $z_{1:T}$, can be performed exactly conditioned on states $\vx_{1:T}$ and observations $\vy_{1:T}$. We first define the forward $\alpha_t$ and backward $\beta_t$ variables as
\begin{align}
	\alpha_t(z_t, c_t) &= p( \vy_{1:t}, \vx_{1:t}, z_t, c_t),\\
	\beta_t(z_t, c_t) &= p(\vy_{t+1:T}, \vx_{t+1:T} | \vx_t, z_t, c_t).
\end{align} 
The smoothed posterior over the switches and counts, $p(z_t, c_t |  \vy_{1:T}, \vx_{1:T})$, can be computed using $\alpha_t$ and $\beta_t$ as 
\begin{equation}
\begin{aligned}
	\gamma_t(z_t, c_t) = p(z_t, c_t |  \vy_{1:T}, \vx_{1:T}) &\propto  p( \vy_{1:T}, \vx_{1:T}, z_t, c_t)\label{eq:gamma-posterior}\\
	&= p(\vy_{t+1:T}, \vx_{t+1:T} |  \vx_{t}, z_t, c_t) p(\vy_{1:t}, \vx_{1:t}, z_t, c_t)\\
	&= \beta_t(z_t, c_t)\alpha_t(z_t, c_t).
\end{aligned}
\end{equation}

In the following, we derive the recursions for $\alpha_t$ and $\beta_t$, similar to the forward-backward algorithm for HSMMs~\cite[Chapter 3]{chiappa2019explicit}:
\begin{equation}
\begin{aligned}
	\alpha_1(z_1, c_1) &= \delta_{c_1 = 1} p(\vy_1, \vx_1| z_1) p(z_1),\\
	\alpha_t(z_t, c_t) &= p(\vy_{1:t}, \vx_{1:t}, z_t, c_t)\\
	&= \sum_{z_{t-1}} \sum_{c_{t-1}} p(\vy_{1:t}, \vx_{1:t}, z_{t-1}, z_t, c_{t-1}, c_t)\\
	&=\sum_{z_{t-1}} \sum_{c_{t-1}} \Big[ p(\vy_t, \vx_t|\vx_{t-1}, z_t)p(z_t| \vx_{t-1}, z_{t-1}, c_t) \\
	& \hspace{19mm} \cdot p(c_t | z_{t-1}, c_{t-1}) p(\vy_{1:t-1}, \vx_{1:t-1}, z_{t-1}, c_{t-1})\Big]\\
	&= \sum_{z_{t-1}} \sum_{c_{t-1}} p(\vy_t, \vx_t|\vx_{t-1}, z_t)p(z_t | \vx_{t-1}, z_{t-1}, c_t) p(c_t | z_{t-1}, c_{t-1}) \alpha(z_{t-1}, c_{t-1})\\
	&= 
	p(\vy_t, \vx_t|\vx_{t-1}, z_t)\Bigg[\delta_{
		\begin{subarray}{l}z_{t-1}=z_t
			\\ c_t > 1
			\\ c_{t-1}=c_t-1
	\end{subarray}} \!\! v_{z_{t-1}}(c_{t-1}) \alpha(z_{t-1}, c_{t-1})\nonumber\\
	&\quad+ \delta_{c_t = 1}\sum_{z_{t-1}}p(z_t |  \vx_{t-1}, z_{t-1}, c_t)\sum_{c_{t-1}}(1-v_{z_{t-1}}(c_{t-1})) \alpha(z_{t-1}, c_{t-1})\Bigg],
\end{aligned}
\end{equation}

\begin{equation}
\begin{aligned}
	\beta_T(z_T, c_T) & = 1,\\
	\beta_t(z_t, c_t) &= p(\vy_{t+1:T}, \vx_{t+1:T} | \vx_t, z_t, c_t)\\
	&= \sum_{z_{t+1}, c_{t+1}} p(\vy_{t+1:T}, \vx_{t+1:T}, z_{t+1}, c_{t+1} | \vx_t, z_t, c_t)\\
	&= \sum_{z_{t+1}, c_{t+1}} \Bigg[p(\vy_{t+2:T}, \vx_{t+2:T}| \vx_{t+1}, z_{t+1}, c_{t+1})p(\vy_{t+1}, \vx_{t+1}|\vx_{t}, z_{t+1})\\
	&\hspace{20mm} \cdot p(z_{t+1}|\vx_t, z_t, c_{t+1})p(c_{t+1}|z_t, c_t ) \Bigg]\\
	&= 	\delta_{\begin{subarray}{l}z_{t+1} = z_t\\c_{t+1} = c_t +1\end{subarray}}\beta_{t+1}(z_{t+1}, c_{t+1})p(\vy_{t+1}, \vx_{t+1}|\vx_{t}, z_{t+1})v_{z_{t}}(c_{t})\nonumber\\
	&\quad + \delta_{\begin{subarray}{l}c_{t} \geq \dmin\\c_{t+1} = 1\end{subarray}}(1-v_{z_{t}}(c_{t})) \!\! \sum_{z_{t+1}} \! \beta_{t+1}(z_{t+1}, c_{t+1})p(\vy_{t+1}, \vx_{t+1}|\vx_{t}, z_{t+1})  p(z_{t+1}|  \vx_t, z_t, c_{t+1}).
\end{aligned}
\end{equation}

\subsection{RED-SDS instantiation}
\label{subsec:instantiation}
In this section, we present our instantiation of RED-SDS, providing the general structure of the architecture and the functions $f_z$, $f_x^\mu$, $f_x^\Sigma$, $f_y^\mu$, $f_y^\Sigma$ (cf. Section \ref{subsec:model-formulation}). For specific implementation details we refer the reader to Appendices \ref{app:seg-hyperparam} and \ref{app:forecast-hyperparam} for segmentation and forecasting, respectively.

\paragraph{Control embedding network.} Control inputs can be either static, i.e., constant for all timesteps per time series (e.g., time series ID), or time-dependent (e.g., time embedding or other covariates). We denote the raw control input as $\vu_{\text{raw}} = [\vu^{\text{static}} ,\vu^{\text{time}}]$, where $\vu^{\text{static}}$ and $\vu^{\text{time}}$ correspond to the static and time-dependent control features, respectively.

In our forecasting experiments, we consider static features that are natural numbers (corresponding to time series IDs).
The static features are first passed through an embedding layer. %
The embedding of the static features is then concatenated with the time-dependent features and the result is fed into a single hidden layer MLP that outputs the control input $\vu_t \in \mathbb{R}^{c}$. This process is described by the following operations:
\begin{equation}
\label{eq:control_net}
\vu_t = f_u\left(g^u_{\mathrm{emb}}(\vu_t^{\text{static}}), \vu_t^{\text{time}}\right),
\end{equation}
where $g^u_{\mathrm{emb}}$ denotes the embedding layer and $f_u$ is the MLP.

\paragraph{Inference network.} The observations $\vy_{1:T}$ are first passed through an embedding network $g^y_{\mathrm{emb}}$, that can either be a bi-RNN or a Transformer, which outputs an embedding $\mathbf{h}^1_{1:T}$ of the observations.
The embedding is then fed into a causal RNN $g_{\mathrm{rnn}}$ along with the control inputs $\vu_{1:T}$ which outputs hidden states $\mathbf{r}_{1:T}$. A single hidden layer MLP $g_{\mathrm{fc}}$ then maps the hidden states $\mathbf{r}_{1:T}$ to the parameters of the Gaussian distribution $q(\vx_{t}|\vx_{1:t-1}, \mathbf{h}^1_{1:T})$. This inference network is described by the following operations:
\begin{align}
	\mathbf{h}^1_{1:T} &= g^y_{\mathrm{emb}}(\vy_{1:T}),\label{eq:infer-embed}\\
	\mathbf{r}_t &= g_{\mathrm{rnn}}( \vx_{t-1}, \mathbf{r}_{t-1}, \vu_{t}, \mathbf{h}^1_{t}),\\
	\vx_t &\sim \mathcal{N}\left(\vx_t; g^\mu_{\mathrm{fc}}(\mathbf{r}_t), g^\Sigma_{\mathrm{fc}}(\mathbf{r}_t)\right),
\end{align}  
where $\vx_0$ and $\mathbf{r}_0$ are zero vectors. 

\paragraph{Duration network.} When no control input is available, the duration model is a learnable matrix $\vP\in\bR^{K\times (\dmax - \dmin + 1)}$, fixed across all timesteps; the $k$-th row represents the logits of the duration model $\rho_k$ for the $k$-th switch. When control inputs are used, the duration model depends on the control input and is no longer fixed across timesteps. The control input $\vu_t$ (see Eq.\ \ref{eq:control_net}) is fed into a single hidden layer MLP which outputs a (time-varying) matrix $\vP_t\in\bR^{K\times \dmax}$, i.e.,
\begin{equation}
	\vP_t = f_d(\vu_t).
\end{equation}
where $\vP_t$ has the same structure as $\vP$. The first $\dmin-1$ columns are masked and the final duration models $\rho_k$ are obtained by applying the tempered softmax function to the rows of $\vP_t$,
\begin{align}
	\rho_k(d)_t =  \mathrm{Cat}(d; \mathcal{S}_{\tau_\rho}(\vP_t^{k, :})),
\end{align}
where $\vP_t^{k, :}$ denotes the $k$-th row of $\vP_t$ and $\mathcal{S}_{\tau_\rho}$ denotes the tempered softmax function with temperature ${\tau_\rho}$.

\paragraph{Discrete transition network.}

The pseudo-observations of the states $\vx_1, \dots, \vx_{t-1}$, sampled from the inference network, along with the control inputs $\vu_2, \dots, \vu_T$, are passed to the neural network $f_z$ (for the case when $c_t = 1$), a single hidden layer MLP,  to model the discrete transition distribution,
\begin{align}
	p(z_t| \vx_{t-1}, z_{t-1}, c_t, \vu_t) &= 
	\begin{cases}
		\delta_{z_t = z_{t-1}} & \text{if}\quad c_t > 1\\
		\mathrm{Cat}(z_t; \mathcal{S}_{\tau_z}(f_z(\vx_{t-1}, z_{t-1}, \vu_t))) & \text{if}\quad c_t = 1
	\end{cases},
\end{align}
where $\mathcal{S}_{\tau_z}$ denotes the tempered softmax function with temperature ${\tau_z}$. The network $f_z$ takes $\vx_{t-1}$ and $\vu_t$ as input and outputs a matrix $\tilde{\vA}_t \in \mathbb{R}^{K \times K}$. Each row of the matrix $\tilde{\vA}_t \in \mathbb{R}^{K \times K}$ is normalized using $\mathcal{S}_{\tau_z}$ to obtain the stochastic transition matrix $\vA_t$ where each row represents a categorical distribution that can be indexed by $z_{t-1}$.

\paragraph{Continuous transition network.} The continuous transition network $f_x$ is a linear function or a single hidden layer MLP that models the continuous transition distribution
\begin{align}
	p(\vx_t|\vx_{t-1}, z_t, \vu_t) &=  \mathcal{N}\left(\vx_t; f_x^\mu(\vx_{t-1}, z_t, \vu_t), f_x^\Sigma(\vx_{t-1}, z_t, \vu_t)\right).
\end{align}
The function $f_x$ takes $\vx_{t-1}$ and $\vu_t$ as input and outputs the parameters of the Gaussian distribution. The dependence on $z_t$ is realized by using separate functions $f_x^k$ for the $K$ unique values of the switch $z_t$.

\paragraph{Emission network.}
The emission network $f_y$ is a linear function or an MLP with two hidden layers that models the emission distribution
\begin{align}
	p(\vy_t|\vx_t) &= \mathcal{N}\left(\vy_t; f_y^\mu(\vx_t), f_y^\Sigma(\vx_t)\right).
\end{align}

\subsection{Applications}
In this section, we describe how to perform time series segmentation and generate probabilistic forecasts using RED-SDS.

\subsubsection{Segmentation}
We perform time series segmentation by labeling every timestep with the most likely switch. This is done by first computing the posterior distribution $\gamma_t(z_t, c_t)$ (Eq. \ref{eq:gamma-posterior}) for each timestep and then obtaining the most likely switch $\hat{k}_t$ by marginalizing out the count variable $c_t$ as follows 
\begin{align}
	\hat{k}_t = \underset{j}{\mathrm{arg max}} \sum_{d = \dmin}^{\dmax}\gamma_t(z_t=j, c_t=d).
\end{align}

\subsubsection{Forecasting}

We generate probabilistic forecasts by generating multiple future sample paths. Let $\vy_{1:T}$ be an input time series and $\tau$ the forecast horizon. We begin by generating $M$ state samples from the variational posterior $q(\vx_T| \vy_{1:T})$ and the corresponding $M$ switch-count pairs from the posterior over the switches and counts $p(z_T, c_T | \vy_{1:T}, \hat{\vx}_{1:T})$. These $M$ triplets of state, switch, and count are then used to unroll the generative model into the future for $\tau$ timesteps generating $M$ sample paths (forecasts). Algorithm \ref{alg:unrolling} describes the steps involved to generate one such sample path.

\makeatletter
\newcommand{\HEADER}[1]{\ALC@it\underline{\textsc{#1}}\begin{ALC@g}}
	\newcommand{\ENDHEADER}{\end{ALC@g}}
\makeatother

\begin{algorithm}
	\caption{RED-SDS future unrolling}
	\label{alg:unrolling}
	\begin{algorithmic}
		
		\HEADER{Input}
		\STATE Time series $\vy_{1:T}$, forecast horizon $\tau$ 
		\ENDHEADER
		\vspace{1mm}
		
		\HEADER{Sample states at $T$}
		\STATE Continuous state sample from the variational posterior $\hat{\vx}_T \sim q(\vx_T| \vy_{1:T})$\vspace{0.5mm}
		\STATE Discrete state and duration samples $\hat{z}_T, \hat{c}_T \sim p(z_T, c_T | \vy_{1:T}, \hat{\vx}_{1:T}) = \gamma_T(z_T, c_T)$, where $\gamma_T(z_T, c_T)$ is computed using the forward-backward algorithm \vspace{0.5mm}
		\STATE Set $\hat{\vy}_{T}=\vy_{T}$
		\ENDHEADER
		\vspace{1mm}
		
		\HEADER{Unroll in forecast horizon}
		\FOR{$t=T+1:T+\tau$ }\vspace{0.5mm}
		\STATE $\hat{c}_{t} \sim p(c_{t} |  \hat{z}_{t-1}, \hat{c}_{t-1})$\vspace{0.5mm}
		\STATE $\hat{z}_{t} \sim p(z_{t} |  \hat{\vx}_{t-1}, \hat{z}_{t-1}, \hat{c}_{t})$\vspace{0.5mm}
		\STATE $\hat{\vx}_{t} \sim p(\vx_{t} | \hat{\vx}_{t-1}, \hat{z}_{t} )$\vspace{0.5mm}
		\STATE $\hat{\vy}_{t} \sim p(\vy_{t} | \hat{\vx}_{t})$\vspace{0.5mm}
		\ENDFOR
		\ENDHEADER
		\vspace{1mm}
		
		\HEADER{Return}
		\STATE Predictive samples $\hat{\vy}_{T+1:T+\tau}$
		\ENDHEADER

	\end{algorithmic}
\end{algorithm}

\section{Details on segmentation experiments}
\label{app:exp_seg}

\subsection{Datasets}
\label{app:seg_datasets}

\paragraph{Bouncing ball.} The bouncing ball dataset comprises univariate time series $y_{1:T}$, with $y_{t}\in\bR$, that encode the location of a ball bouncing between two fixed walls with constant absolute velocity and elastic collisions between the ball and the wall(s). The distance between the walls was set to 10 with the initial location of the ball randomly generated between the walls. The initial velocity was sampled from the uniform distribution $\mathcal{U}(-0.5, 0.5)$ and its sign was flipped every time the ball went beyond 0 or 10. The final observation was generated by adding $\epsilon \sim \mathcal{N}(0, 0.1^2)$ to the ball's location. The ground truth label was assigned based on the sign of the velocity. We generated 100000 and 1000 time series of 100 timesteps for the train and the test datasets respectively.

\paragraph{3 mode system.} The 3 mode system dataset comprises univariate time series generated from a switching linear dynamical system with 3 modes and an explicit duration model for each mode. The dimensionality of the state variables $\vx_t$ was set to 2. The initial switch and state distributions were given by
\begin{align}
	p(z_1) &= \mathcal{U}_{\text{cat}}\{1, 2, 3\},\\
	p(\vx_1 | z_1) &= \mathcal{N}\left(\begin{bmatrix}
		2 \\ 0
	\end{bmatrix}, 0.01\mathbf{I}\right),
\end{align}
where $\mathcal{U}_{\text{cat}}$ denotes the uniform categorical distribution.

The transition distribution of the increasing count variables was given by
\begin{align}
	p(c_t | z_{t-1} = k, c_{t-1} ) &= 
	\begin{cases}
		v_{k}(c_{t-1}) & \text{if}\quad c_t = c_{t-1} + 1\\
		1 - v_{k}(c_{t-1}) & \text{if}\quad c_t = 1\\
	\end{cases},
\end{align}
where the probability of a count increment $v_k$ for the switch $k$ is defined as in Eq. \eqref{eq:count_incr} via the duration models $\rho_k$, given by
{\footnotesize
	\begin{align}
		\rho_1(d) &= \mathrm{Cat}\left(\begin{bmatrix} \frac{2}{17} & 0 & 0 & 0 & 0 & \frac{5}{17} & 0 & 0 & 0 & 0 & \frac{7}{17} & 0 & 0 & 0 & \frac{3}{17} \end{bmatrix}\right),\\
		\rho_2(d) &= \mathrm{Cat}\left(\begin{bmatrix} 0 & 0 & \frac{1}{4} & 0 & 0 & 0 & 0 & 0 & 0 & 0 & 0 & \frac{2}{5} & 0 & \frac{3}{10} & \frac{1}{20} \end{bmatrix}\right),\\
		\rho_3(d) &= \mathrm{Cat}\left(\begin{bmatrix} 0 & 0 & 0 & 0 & 0 & 0 & 0 & \frac{3}{17} & 0 & 0 & \frac{7}{17} & 0 & \frac{5}{17} & 0 & \frac{2}{17} \end{bmatrix}\right),
	\end{align}
}%
with $\dmin = 6$ and $\dmax = 20$.

The switch transition distribution was given by
\begin{align}
	p(z_t = j| z_{t-1} = i, c_t) &= 
	\begin{cases}
		\delta_{z_t = i} & \text{if}\quad c_t > 1\\
		\begin{bmatrix} 0.1 & 0.2 & 0.7\\
			0.3 & 0.5 & 0.2\\
			0.3 & 0.3 & 0.4\\ \end{bmatrix}_{(i, j)} & \text{if}\quad c_t = 1
	\end{cases}.
\end{align}

The state transition distribution was given by
\begin{align}
	p(\vx_{t} | \vx_{t-1}, z_t = k) &= \mathcal{N}(\mathbf{A}_k\vx_{t-1} + \mathbf{b}_k, 0.01\mathbf{I}),
\end{align}
where the $\mathbf{A}$s and $\mathbf{b}$s were defined as follows
\begin{align}
	\mathbf{A}_1 &= 0.99\begin{bmatrix}
		\cos(0) & -\sin(0)\\
		\sin(0) & \cos(0)\\
	\end{bmatrix},\\
   \mathbf{b}_1 &= \bm{0},\\
	\mathbf{A}_2 &= 0.99\begin{bmatrix}
		\cos(\pi/8) & -\sin(\pi/8)\\
		\sin(\pi/8) & \cos(\pi/8)\\
	\end{bmatrix},\\
	\mathbf{b}_2 &= 0.25\bm{\epsilon}_2,\\
	\mathbf{A}_3 &= 0.99\begin{bmatrix}
		\cos(\pi/4) & -\sin(\pi/4)\\
		\sin(\pi/4) & \cos(\pi/4)\\
	\end{bmatrix},\\
	\mathbf{b}_3 &=0.25\bm{\epsilon}_3,
\end{align}
with $\bm{\epsilon}_2$, $\bm{\epsilon}_3$ sampled from $\mathcal{N}(\bm{0}, \mathbf{I})$.

The emission distribution was given by 
\begin{align}
	p(\vy_t | \vx_t, z_t=k) &= \mathcal{N}\left(\mathbf{c}^\top_k\vx_{t} + d_k, 0.04\mathbf{I}\right),
\end{align}
with $\mathbf{c}_k \sim \mathcal{N}(\bm{0}, \mathbf{I})$ and $d_k \sim \mathcal{U}_{\text{cat}}\{0, 1, 2\}$.

The ground truth label was assigned based on $z_t$. We generated 10000 and 500 time series of 180 timesteps for the train and the test datasets, respectively.

\paragraph{Dancing bees.} We used the publicly-available\footnote{\url{https://www.cc.gatech.edu/~borg/ijcv_psslds/}} dancing bees dataset~\cite{oh2008learning}, which comprises tracks of six dancer honey bees that were obtained using a vision-based tracker. The time series consist of the 2D coordinates $(x_t, y_t)$ and the heading angle $(\theta_t)$ of the honey bee at every timestep. Each timestep is labeled as one of the three dance types: waggle, turn right, and turn left. The six long time series have lengths 1058, 1125, 1054, 757, 609 and 814. We first standardized the 2D coordinates $(x_t, y_t)$ for each time series by subtracting the mean and dividing by the standard deviation to get the normalized coordinates $(\hat{x}_t, \hat{y}_t)$. We then constructed 4-dimensional time series with elements $[\hat{x}_t, \hat{y}_t, \sin(\theta_t), \cos(\theta_t)]$. Each time series was split into chunks of 120 timesteps starting at every changepoint in the ground truth label (discarding terminal chunks less than 120 timesteps). We used time series $\{1, 3, 4, 5, 6\}$ for training and the longest time series $\{2\}$ for testing, resulting in 84 and 21 data points in the train and test datasets, respectively.

\subsection{Segmentation metrics}
\label{app:seg-metrics}
The segment label ordering assigned by a model such as RED-SDS---trained without label supervision---is arbitrary and may not match with the ground truth label ordering. Thus, we evaluated the segmentation performance using three metrics that are independent of remappings of the predicted labels:
\begin{enumerate}[label=(\alph*)]
	\item Frame-wise segmentation accuracy: This metric computes the percentage of predicted labels that match the ground truth labels. The predicted labels were first remapped using the Hungarian algorithm~\cite{kuhn1955hungarian} which finds the mapping that maximizes the accuracy. We used \texttt{linear\_sum\_assignment} from \texttt{scipy} to find the optimal mapping.
	\item Normalized Mutual Information (NMI)~\cite{vinh2010information}: NMI computes the mutual information between two labelings, normalized to lie between 0 (no mutual information) and 1 (perfect correlation). The metric is independent of the actual values of labels, i.e., a remapping of the labels won't change the score. We used \texttt{normalized\_mutual\_info\_score} from \texttt{scikit-learn} to compute the NMI score.
	\item Adjusted Rand Index (ARI)~\cite{hubert1985comparing}: ARI computes a similarity measure between two labelings which is adjusted for chance. The metric lies between -1 to 1 where a random labeling has a score close to 0 and a score of 1 indicates a perfect match. Like NMI, ARI is independent of the actual values of the labels. We used \texttt{adjusted\_rand\_score} from \texttt{scikit-learn} to compute the ARI score.
\end{enumerate}

\subsection{Training and hyperparameter details}
\label{app:seg-hyperparam}
In this section, we discuss the training and hyperparameter details for the segmentation experiments.

\paragraph{Training parameters.} We trained all the datasets with a fixed batch size of 32 for 20000 training steps. We used the Adam optimizer for the gradient updates with $10^{-5}$ weight-decay and clipped the gradient norm to 10. The learning rate was warmed-up linearly from a lower value in $\{5\times10^{-5}, 1\times10^{-4}\}$ to $\{2\times10^{-4}, 5\times10^{-3}\}$ for the first $\{2000, 1000\}$ steps after which a cosine decay follows for the remaining time steps with a decay rate of 0.99.

\begin{table*}[htb]
	\footnotesize
	\centering
	\caption{\small Network architectures for the different components of the model. \texttt{Linear} denotes a linear layer without bias, \texttt{MLP} [$a_1$, \dots, $a_l$] denotes an $l$-hidden-layer MLP with hidden units $a_1$, \dots, $a_l$ and ReLU non-linearity, \texttt{biGRU} [$b$] denotes a single-layer bidirectional GRU with $b$ hidden units, and \texttt{RNN} [$c$] denotes a single-layer RNN with $c$ hidden units.}
	\label{tab:seg_arch}
	\begin{tabular}{lccc}
		\toprule
		\multirow{2}{*}{Network} & \multicolumn{3}{c}{Datasets} \\
		\cmidrule{2-4}
		&      \bb &        \edslds & \bee  \\
		\midrule
		Discrete Transition ($f_z$) & \texttt{MLP} [$4 \times 2^2$] & \texttt{MLP} [$4 \times 3^2$] & \texttt{MLP} [$4 \times 3^2$]\\
		Continuous Transition ($f_x$) & \texttt{MLP} [32] & \texttt{MLP} [32] & \texttt{Linear}\\
		Emission Network ($f_y$) & \texttt{MLP} [8, 32] & \texttt{MLP} [8, 32] & \texttt{Linear}\\
		\midrule
		Inference Embedder ($g^y_{\mathrm{emb}}$) & \texttt{biGRU} [4] & \texttt{biGRU} [4]  & \texttt{biGRU} [16]\\
		Causal RNN ($g_{\mathrm{rnn}}$) & \texttt{RNN} [16] & \texttt{RNN} [16] & \texttt{RNN} [16]\\
		Parameter Network ($g_{\mathrm{fc}}$) & \texttt{MLP} [32] & \texttt{MLP} [32] & \texttt{MLP} [32] \\
		\bottomrule
	\end{tabular}
\end{table*}

\begin{table*}[htb]
	\footnotesize
	\centering
	\caption{\small Temperature annealing details.}
	\label{tab:seg_anneal}
	\subfloat[Annealing schedules of the switch ($\tau_z$) and duration ($\tau_\rho$) temperatures.]{
	\begin{tabular}{lccc}
		\toprule
		\multirow{2}{*}{Hyperparameter} & \multicolumn{3}{c}{Datasets} \\
		\cmidrule{2-4}
		&      \bb &        \edslds & \bee  \\
		\midrule
        Initial Temp. & $\tau_z, \tau_\rho$: 10 & $\tau_z, \tau_\rho$: 10 & $\tau_z$: 100, $\tau_\rho$: 10 \\
		Min Temp. & $\tau_z, \tau_\rho$: 1 & $\tau_z, \tau_\rho$: 1 & $\tau_z$: 10, $\tau_\rho$: 1 \\
		Decay Rate & 0.99 & 0.99 & 0.95 \\
		Begin Decay Step & 1000 & 1000 & 5000 \\
		Decay Every (steps) & 50 & 50 & 100 \\
		\bottomrule
	\end{tabular}}\\
	\subfloat[\cmark\ indicates that the temperature is annealed using the schedule in (a) and \xmark\ indicates that the temperature is kept fixed.]{
	\begin{tabular}{llccc}
		\toprule
		\multirow{2}{*}{Model} & \multirow{2}{*}{Variable} & \multicolumn{3}{c}{Datasets} \\
		\cmidrule{3-5}
		& &       \bb &        \edslds & \bee  \\
		\midrule
		\SNLDS & Switch & \cmark & \cmark & \cmark\\
		\cmidrule{2-5}
		\multirow{2}{*}{\EDSDS} & Switch & \xmark & \xmark & \cmark\\
		& Duration & \cmark & \cmark & \cmark\\
		\cmidrule{2-5}
		\multirow{2}{*}{\REDSDS} & Switch & \xmark & \xmark & \cmark\\
		& Duration & \xmark & \cmark & \cmark\\
		\bottomrule
	\end{tabular}
	}
\end{table*}

\paragraph{Network types.} We experimented with linear and non-linear functions for the continuous transition $f_x$ and emission $f_y$ functions. The linear function implies multiplication with a transformation matrix while the non-linear function was an MLP with ReLU non-linearity. For the inference embedding network $g^y_{\mathrm{emb}}$, we used a single layer bidirectional GRU. The causal RNN $g_{\mathrm{rnn}}$ was a single layer forward RNN and the parameter network $g_{\mathrm{fc}}$ was a single hidden layer MLP. Table \ref{tab:seg_arch} summarizes the network architectures of the different model components for the three datasets.

\paragraph{Temperature Annealing.} As discussed in Section \ref{subsec:learning}, we annealed the temperature parameter of the tempered softmax function (Eq. \ref{eq:temp_softmax}) to encourage the model to explore all switches and durations during the initial training phase. We decayed the temperature parameter exponentially from a initial value to a minimum value and report the details on annealing schedules in Table \ref{tab:seg_anneal}.

\paragraph{Duration.} We used the $(\dmin, \dmax)$ hyperparameters to impart prior knowledge to the model about the switch durations. These parameters were set to $(1, 20)$ for the bouncing ball dataset. For the 3 mode system and dancing bees datasets, we used larger values of $(5, 20)$ and $(20, 50)$ respectively to encourage the model to learn longer switch durations as exhibited by these datasets.

\paragraph{Number of switches.} The number of switches $K$ plays an important role in the model, particularly for segmentation. For all the datasets considered in our experiments, the number of operating modes is known; therefore, we set $K$ to the number of ground truth operating modes.

\paragraph{Dimensionality of state $\vx_t$.} The dimensionality of the state variables $\vx_t$ was tuned from the set $\{2, 4\}$ for the univariate datasets (bouncing ball and 3 mode system) and was set to 8 for the multivariate dataset (dancing bees).

\subsection{Baseline models}
\label{app:seg-baselines}
\subsubsection{SNLDS}

In this section, we discuss the objective function of SNLDS proposed by \citet{dong2020collapsed} and the modified version used in our experiments. As mentioned in the main text, SNLDS can be viewed as an ablated version of RED-SDS without the explicit duration models, i.e., with $\dmin = \dmax = 1$. 

The ELBO for SNLDS is given by 
\begin{align}
	\mathrm{ELBO} &= \mathbb{E}_{q(\vx_{1:T}|\vy_{1:T})p(z_{1:T} | \vy_{1:T}, \vx_{1:T})}\left[\log\frac{p( \vy_{1:T}, \vx_{1:T}, z_{1:T})}{q(\vx_{1:T}|\vy_{1:T})p( z_{1:T} | \vy_{1:T}, \vx_{1:T})}\right]\\
	&= \mathbb{E}_{q(\vx_{1:T}|\vy_{1:T})}\left[\log\frac{p( \vy_{1:T}, \vx_{1:T})}{q(\vx_{1:T}|\vy_{1:T})}\right].\label{eq:snlds-elbo}
\end{align}

To estimate the gradients of $\log p(\vy_{1:T}, \vx_{1:T})$ in Eq.~\eqref{eq:snlds-elbo}, \citet{dong2020collapsed} proposed auto-differentiating through the following expression,
\begin{align}
	\mathbb{E}_{p(z_{1:T}| \vy_{1:T}, \vx_{1:T})} \left[\log p(\vy_{1:T}, \vx_{1:T}, z_{1:T})\right] & = \sum_{t=2}^T\sum_{j, k}\xi_t(j, k)\left[\log B_t(k)A_t(j, k)\right]\nonumber\\
	&\quad+ \sum_k \gamma_1(k)\left[\log B_1(k)\pi_k\right],\label{eq:snlds-like-approx}
\end{align}
where
\begin{align}
	\pi(k) &= p(z_1 = k),\\
	\gamma(k) &= p(z_t=k|\vy_{1:T}, \vx_{1:T}),\\
	\xi(j, k) &= p(z_t=k, z_{t-1}=j|\vy_{1:T}, \vx_{1:T}),\\
	B_t(k) &= p(\vy_t|\vx_t)p(\vx_t|\vx_{t-1}, z_t=k),\\
	A_t(j, k) &= p(z_t=j|  \vy_{t-1}, z_{t-1}=k).
\end{align}

However, this makes the model prone to state collapse, i.e., the model ends up only using a single switch, as also noted by \cite{dong2020collapsed}. This led them to add an ad-hoc cross-entropy regularizer to the objective function that discourages state collapse during the initial phase of training. The regularizer minimizes the KL divergence between a uniform prior on the switches and the smoothed posterior over switches,
\begin{align}
	\mathcal{L}_{\mathrm{CE}} &= \sum_{t=1}^T \mathcal{D}_{KL}(p_{\mathrm{prior}}(z_t) \| \gamma(z_t)),\label{eq:snlds-kl-reg}
\end{align}
where $p_{\mathrm{prior}}(z_t) = \prod_j \left[\frac{1}{K}\right]^{\mathbf{1}[j=z_t]}$.

During our initial experiments with SNLDS, we observed frequent state collapse when using the approximation in Eq. \eqref{eq:snlds-like-approx} and the training was sensitive to the annealing schedule of $\mathcal{L}_{\mathrm{CE}}$. We further noted that the likelihood term $p( \vy_{1:T}, \vx_{1:T})$ in Eq. \eqref{eq:snlds-elbo} can be computed exactly (same as in RED-SDS) using the SNLDS forward variable $\alpha_T(z_T) = p( \vy_{1:T}, \vx_{1:T}, z_T)$ as
\begin{align}
	p( \vy_{1:T}, \vx_{1:T}) = \sum_{z_T} \alpha_T(z_T).\label{eq:snlds-exact-likelihood}
\end{align}
Therefore, we used Eq. \eqref{eq:snlds-exact-likelihood} for $p( \vy_{1:T}, \vx_{1:T})$ in our experiments which was less prone to state collapse and didn't require the sensitive KL regularization term (Eq. \ref{eq:snlds-kl-reg}).

\subsubsection{Soft-switching models}

\begin{figure}
	\centering
	\subfloat[{\small Bouncing ball}]{\includegraphics[width=0.5\linewidth]{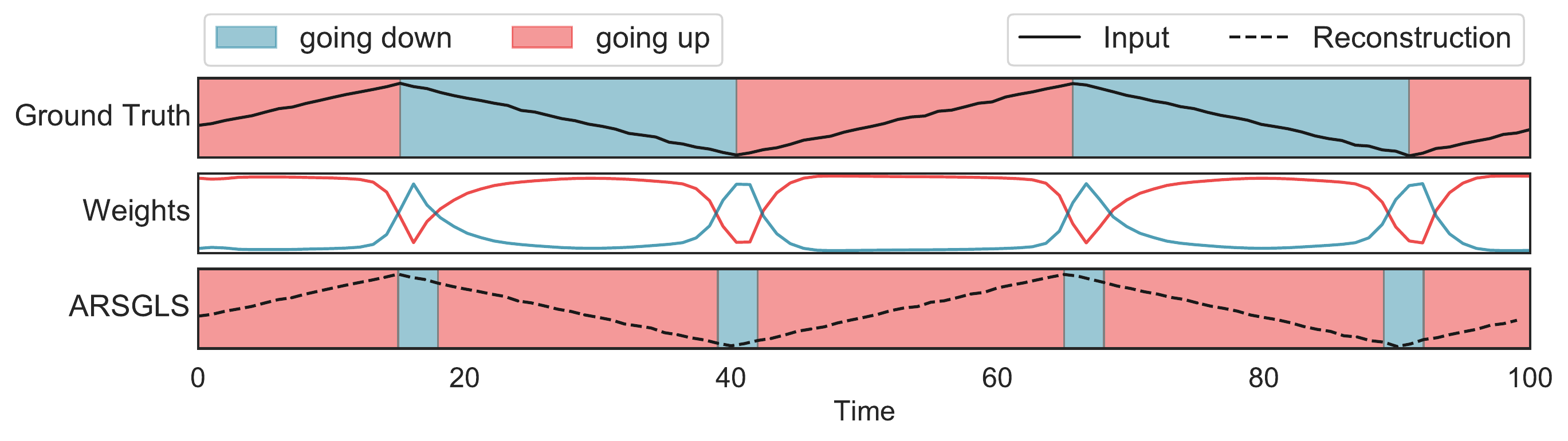}\includegraphics[width=0.5\linewidth]{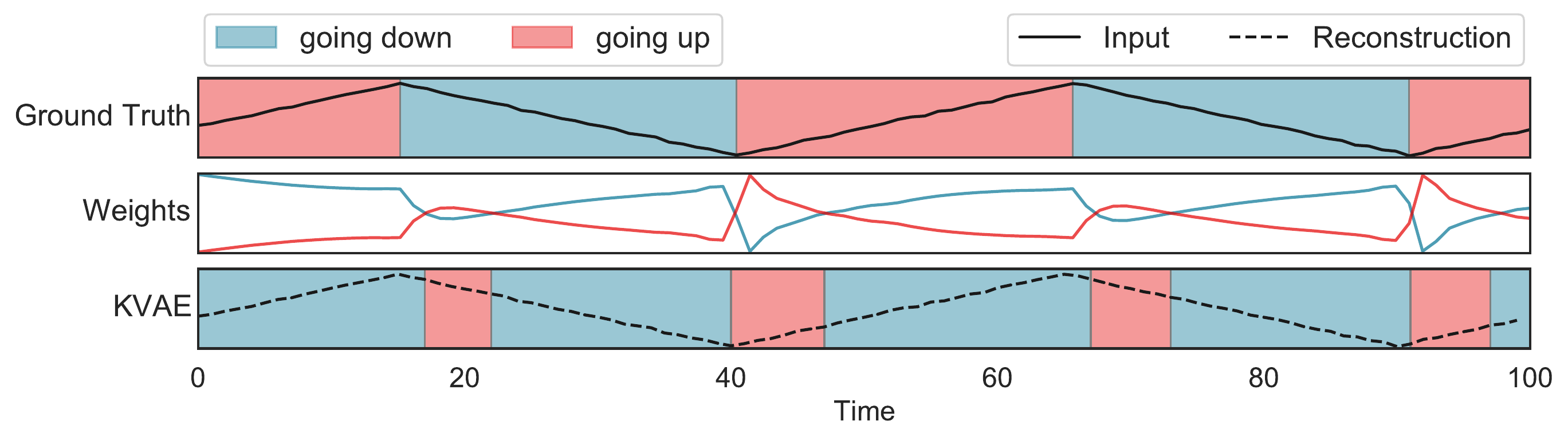}}\\
	\subfloat[3 mode system]{\includegraphics[width=0.5\linewidth]{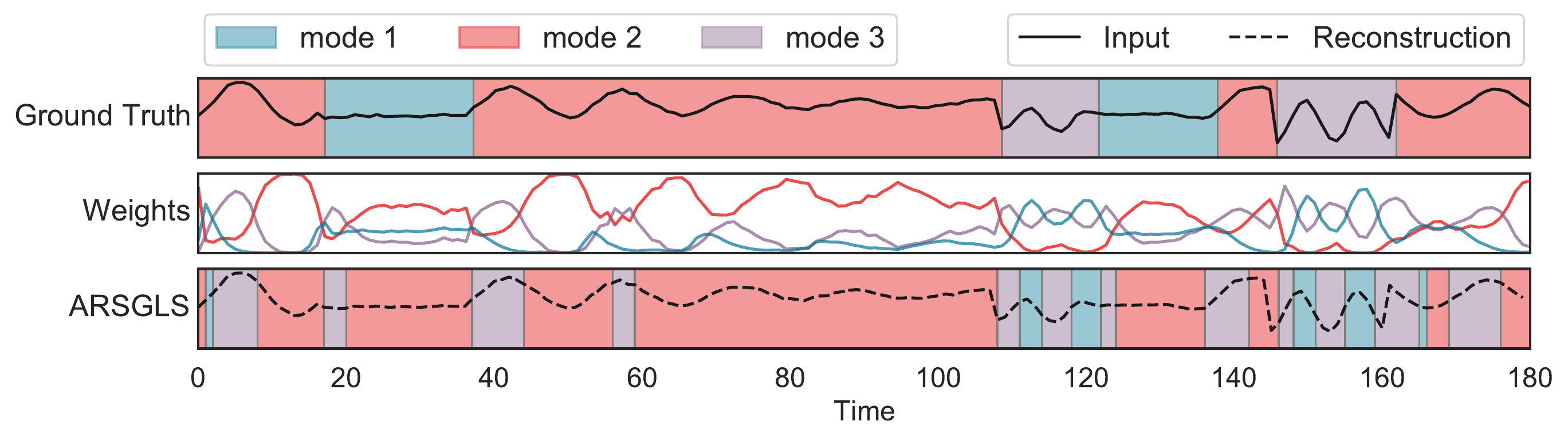}\includegraphics[width=0.5\linewidth]{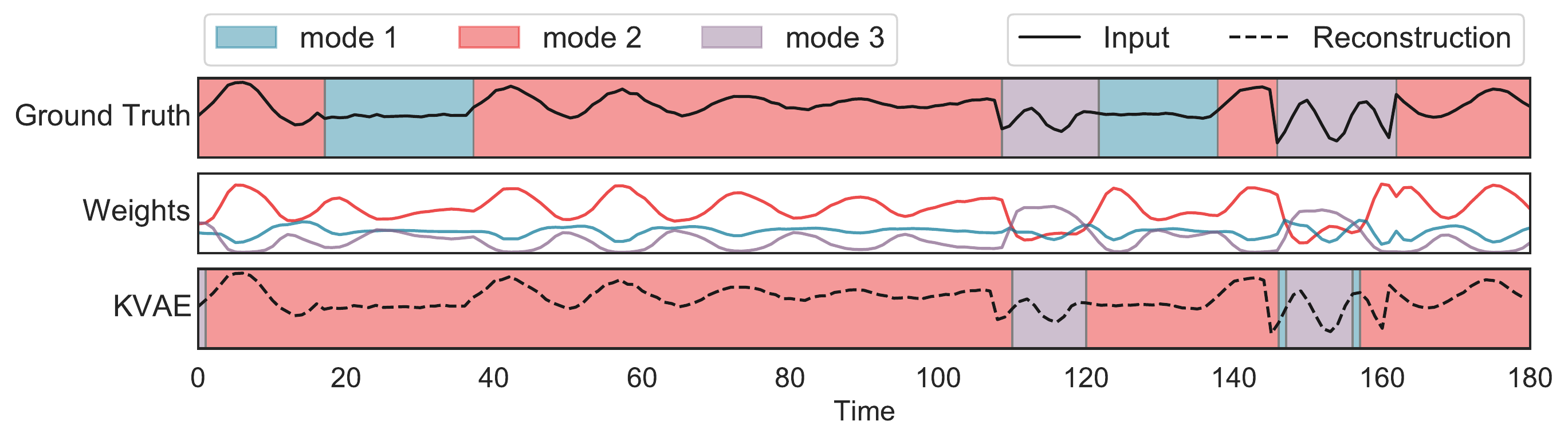}}\\
	\subfloat[Dancing bees]{\includegraphics[width=0.5\linewidth]{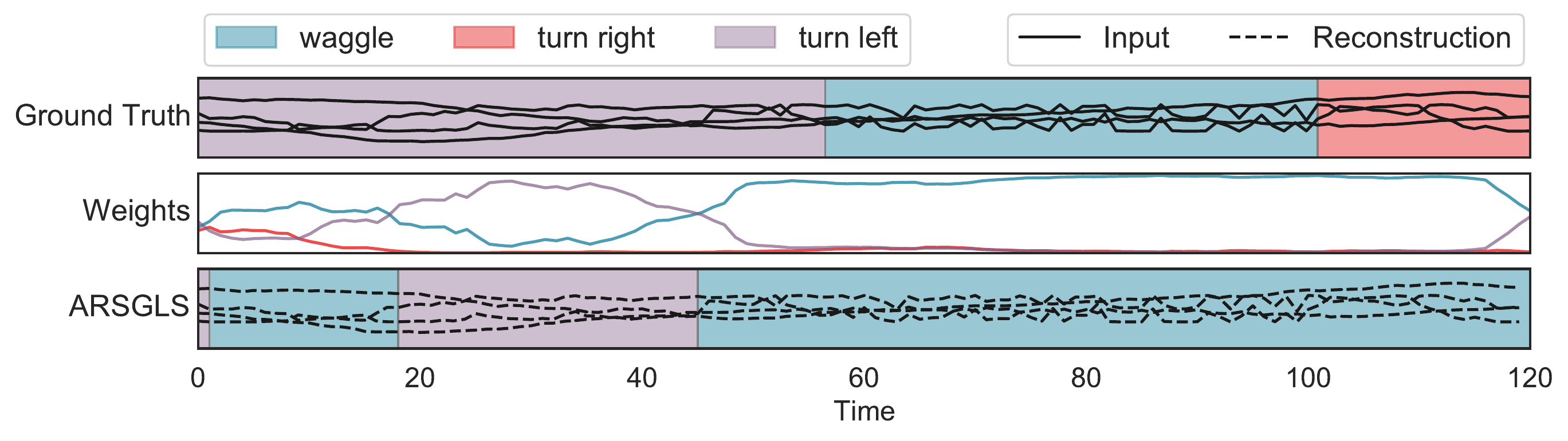}\includegraphics[width=0.5\linewidth]{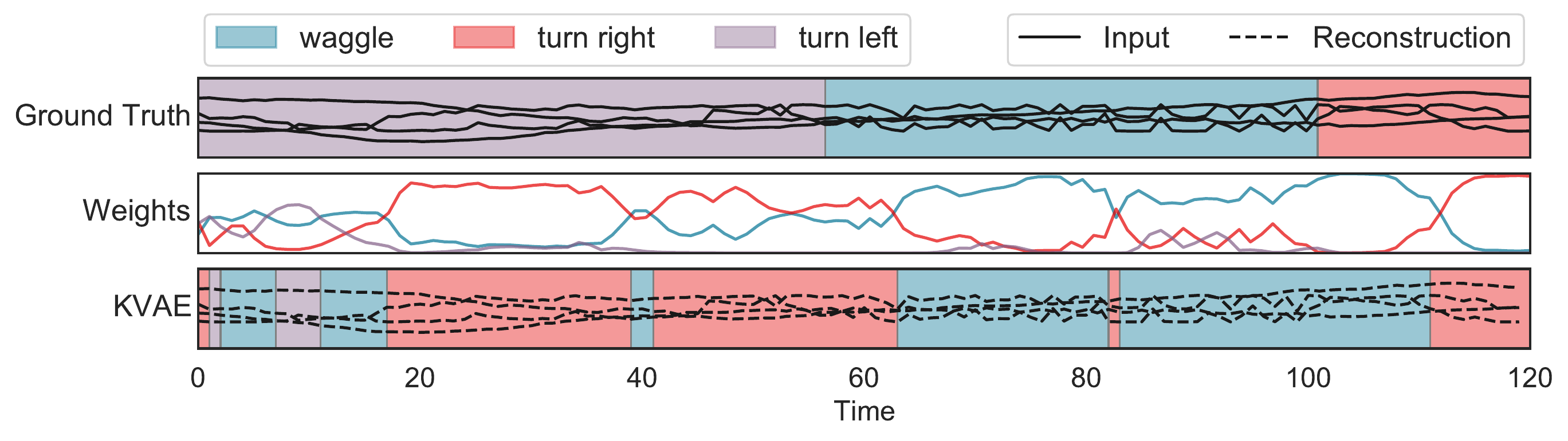}}
	\caption{\small Qualitative segmentation results on the bouncing ball, 3 mode system, and dancing bees datasets for the soft-switching models: ARSGLS (left) and KVAE (right). 
		Background colors represent the different operating modes.
	}
	\label{fig:soft-seg}
\end{figure}

We conducted experiments using two soft-switching models: ARSGLS  and KVAE. The ARSGLS extends the classical SLDS by conditionally linear state-to-switch connections and an auxiliary variable along with a neural network (decoder) that allows it to model more complex non-linear emission models and multivariate observations. 
However, the switch variables in ARSGLS are Gaussian rather than categorical, such that the recurrent state-to-switch connection does not break the computational tractability of the conditionally linear dynamical system. 
Instead of ``indexing'' a distinct base LDS at each timestep, the parameters are predicted as a weighted average of the matrices of the base LDSs.
These combination weights are predicted by a neural network with a softmax transformation which takes the Gaussian switch variables as input.

Similarly, the KVAE uses an LSTM (followed by an MLP) to predict the weights for averaging the base matrices, where the LSTM takes the latent variables embedded by a variational autoencoder as inputs. 

As both ARSGLS and KVAE are trained using a continuous interpolation of operating modes, they cannot always correctly assign a single discrete operating mode to every timestep during test time. Fig. \ref{fig:soft-seg} shows qualitative segmentation results for ARSGLS and KVAE trained on the bouncing ball, 3 mode system, and dancing bees datasets along with the combination weights at each timestep (middle of each plot). The segment labels were assigned by taking the $\arg\max$ of the combination weights. Although the combination weights follow some interpretable patterns (e.g., in the bouncing ball and 3 mode system datasets), they cannot assign each timestep to a single mode. Thus, both ARSGLS and KVAE perform poorly on these segmentation tasks.

\subsection{Additional results}
\label{app:seg-add-results}
Table \ref{tab:extra_results_segmentation} presents segmentation results on the bouncing ball and 3 mode system datasets with the dimenionality of the state $\vx_t$ equal to 2 (ground truth) and 4. All the three models perform similarly for $\mathrm{dim}(\mathbf{x}_t) = 2$ and $\mathrm{dim}(\mathbf{x}_t) = 4$ on the bouncing ball dataset. On the 3 mode system dataset, RED-SDS performs better with $\mathrm{dim}(\mathbf{x}_t) = 4$ than $\mathrm{dim}(\mathbf{x}_t) = 2$.

\begin{figure}[htb]
		\centering
		\subfloat[Bouncing ball]{\includegraphics[width=0.5\linewidth]{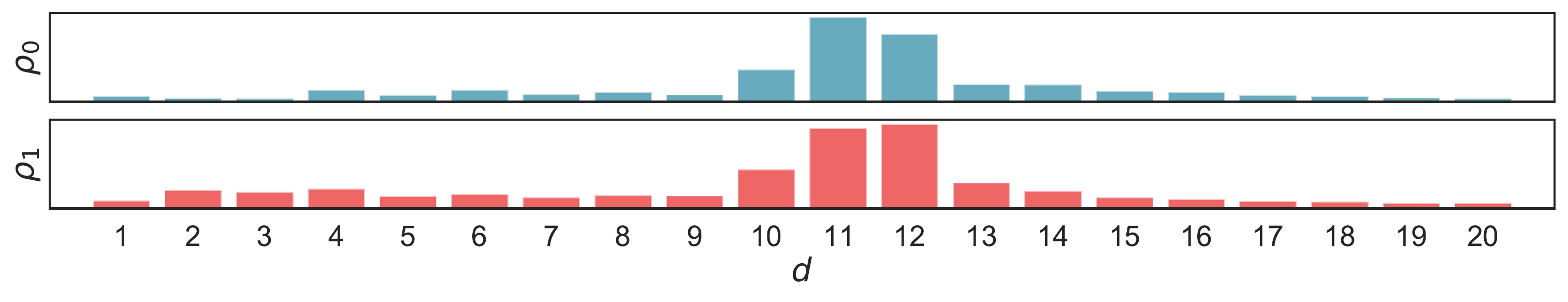}\includegraphics[width=0.5\linewidth]{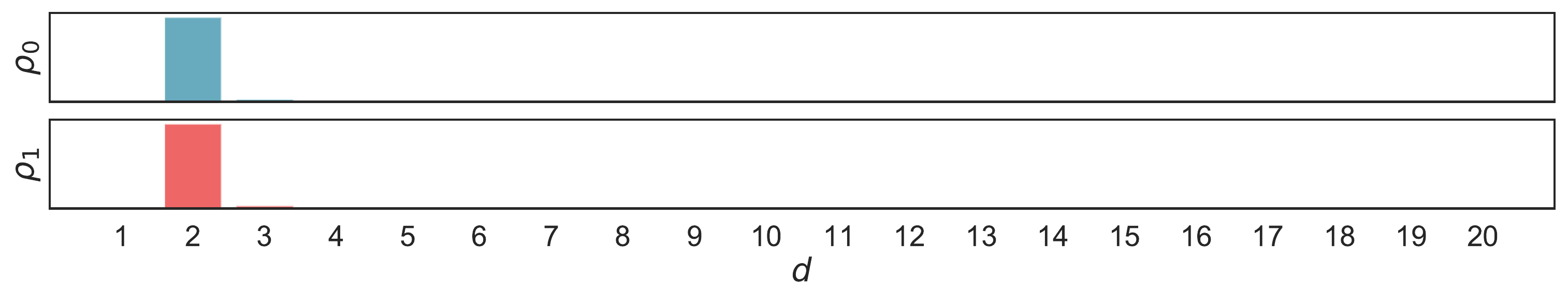}}\\
		\subfloat[3 mode system]{\includegraphics[width=0.5\linewidth]{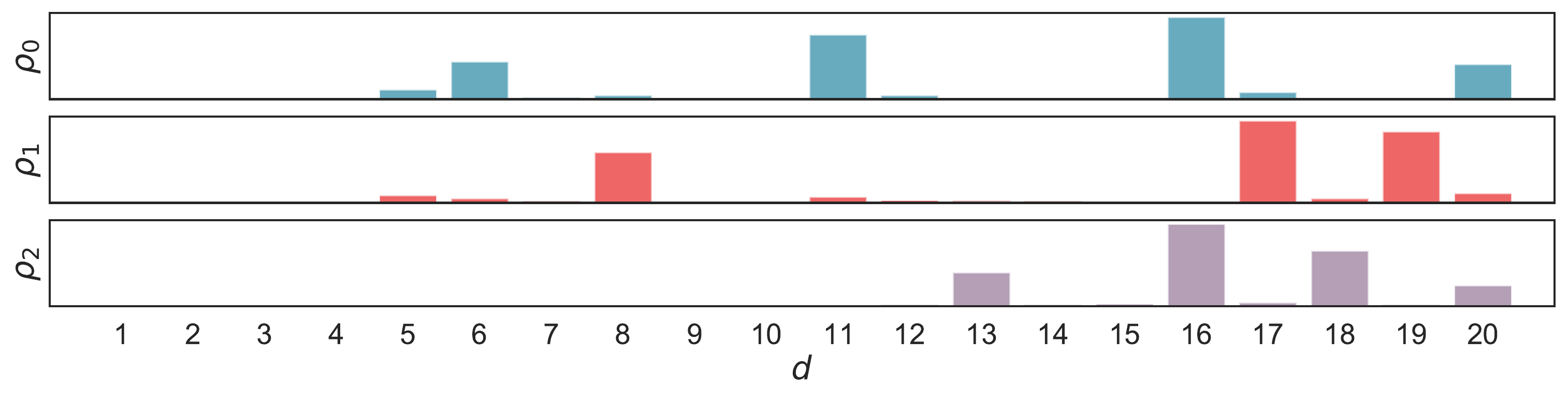}\includegraphics[width=0.5\linewidth]{edslds_learned_duration_redsds}}\\
		\subfloat[Dancing bees]{\includegraphics[width=0.5\linewidth]{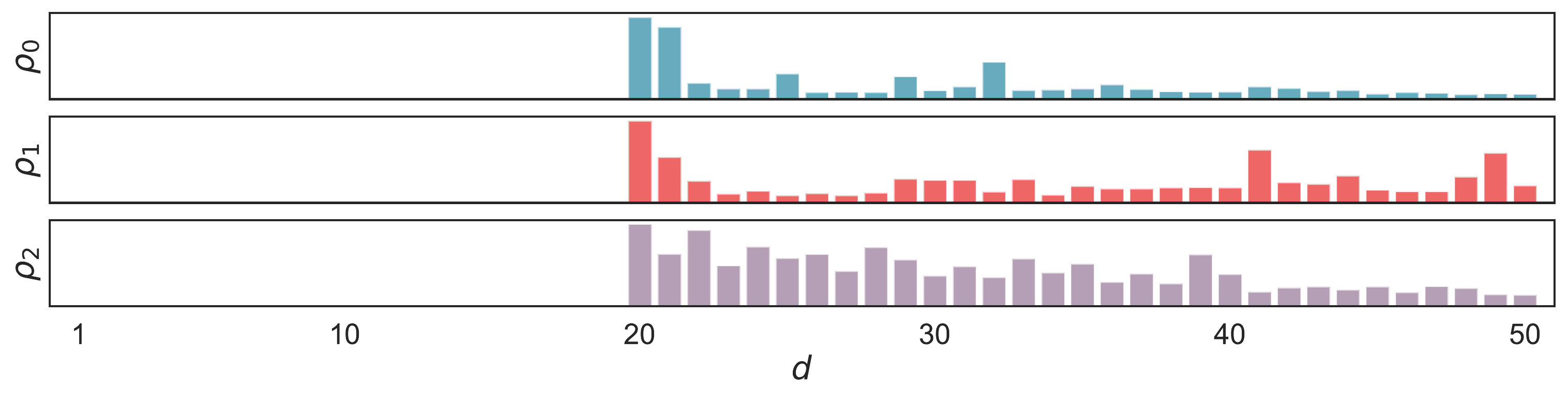}\includegraphics[width=0.5\linewidth]{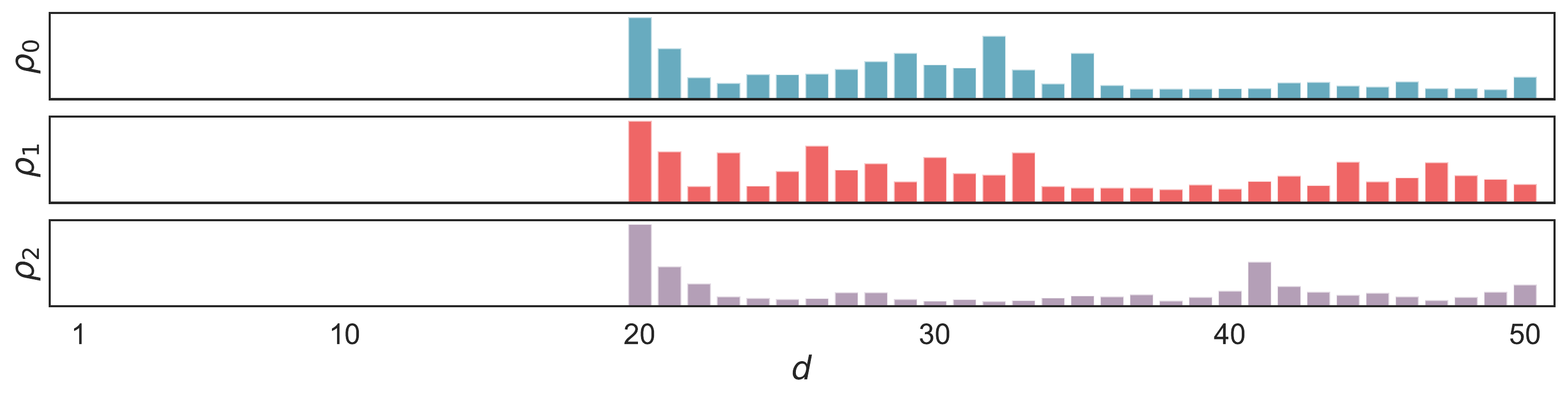}}
		\caption{Duration models learned by ED-SDS (left) and RED-SDS (right) for the bouncing ball, 3 mode system, and dancing bees datasets.}
		\label{fig:extra-duration}
\end{figure}

Fig. \ref{fig:extra-duration} shows the duration models learned by ED-SDS and RED-SDS. On the bouncing ball dataset, RED-SDS assigns all the probability mass to shorter durations indicating that the state-to-switch recurrence is more informative about the switching behavior in this dataset. In contrast, ED-SDS lacks state-to-switch recurrence and assigns probability mass to longer durations. On the 3 mode system dataset, both ED-SDS and RED-SDS recover the ground truth duration model. On the dancing bees dataset, both ED-SDS and RED-SDS assign probability mass to longer durations indicating the existence of long-term temporal patterns.

\begin{table*}[htb]
	\footnotesize
	\centering
	\caption{\small Quantitative results on the bouncing ball and 3 mode system datasets with different values of $\mathrm{dim}(\mathbf{x}_t)$.}
	\begin{tabular}{llccc}
		\toprule
		\multirow{2}{*}{Metric} & \multirow{2}{*}{Model} & \multirow{2}{*}{$\mathrm{dim}(\mathbf{x}_t)$} &     \multicolumn{2}{c}{Datasets} \\
		\cmidrule{4-5}
		& & &      \bb &        \edslds \\
		\midrule
		\multirow{6}{*}{Accuracy} & \multirow{2}{*}{\SNLDS} & 2 &  \textbf{0.97\textpm0.00} & 0.82\textpm0.14 \\
		&  & 4 &  \textbf{0.97\textpm0.00} & 0.82\textpm0.08 \\
		\cmidrule{3-5}
		& \multirow{2}{*}{\EDSDS~(ours)} & 2 & 0.95\textpm0.00 & 0.97\textpm0.00  \\
		&  & 4 & 0.94\textpm0.00 & 0.97\textpm0.00  \\
		\cmidrule{3-5}
		 & \multirow{2}{*}{\REDSDS~(ours)} & 2 & \textbf{0.97\textpm0.00} & 0.95\textpm0.00  \\
		 & & 4 & \textbf{0.97\textpm0.00} &\textbf{ 0.98\textpm0.00}  \\
		\midrule
		\multirow{6}{*}{NMI} & \multirow{2}{*}{\SNLDS} & 2 & \textbf{0.83\textpm0.01} &  0.63\textpm0.12   \\
		&  & 4 & 0.82\textpm0.01 &  0.63\textpm0.08   \\
		\cmidrule{3-5}
		& \multirow{2}{*}{\EDSDS~(ours)} & 2 & 0.71\textpm0.00 & 0.89\textpm0.00  \\
		&  & 4 & 0.70\textpm0.01 & 0.88\textpm0.01  \\
		\cmidrule{3-5}
		 & \multirow{2}{*}{\REDSDS~(ours)} & 2 &  0.80\textpm0.00 & 0.82\textpm0.01  \\
		 &  & 4 &  0.81\textpm0.00 & \textbf{ 0.91\textpm0.01}  \\
		\midrule
		\multirow{6}{*}{ARI} & \multirow{2}{*}{\SNLDS} & 2 &  \textbf{0.90\textpm0.01} & 0.67\textpm0.17 \\
		&  & 4 &  0.89\textpm0.01 & 0.67\textpm0.11 \\
		\cmidrule{3-5}
		& \multirow{2}{*}{\EDSDS~(ours)} & 2 & 0.81\textpm0.01 & 0.93\textpm0.00 \\
		&  & 4 & 0.79\textpm0.01 & 0.93\textpm0.01 \\
		\cmidrule{3-5}
		 & \multirow{2}{*}{\REDSDS~(ours)} & 2 & 0.88\textpm0.01 & 0.89\textpm0.01 \\
		 &  & 4 & 0.88\textpm0.00 & \textbf{0.95\textpm0.01} \\
		\bottomrule
	\end{tabular}
	\vspace{-1.5em}
	\label{tab:extra_results_segmentation}
\end{table*}

\section{Details on forecasting experiments}
\label{app:forecasting}

\subsection{Datasets}
\label{app:for_datasets}

We evaluated RED-SDS on the following five publicly available datasets, commonly used for evaluation forecasting models: 
\begin{itemize}
\item \exchange: daily exchange rate between 8 currencies as used in \cite{lai2018modeling};
\item  \solar: hourly photo-voltaic production of 137 stations in Alabama State used in \cite{lai2018modeling};
\item \electricity: hourly time series of the electricity consumption of 370 customers \cite{dua2017uci};
\item \traffic: hourly occupancy rate, between 0 and 1, of 963 San Francisco car lanes \cite{dua2017uci};
\item \wiki: daily page views of 2000 Wikipedia pages used in \cite{gasthaus2019probabilistic}. 
\end{itemize}
Similar to \cite{kurle2020deeprao}, the forecasts of different methods are evaluated by splitting each dataset in the following fashion: all data prior to a fixed \textit{forecast start date} comprise the training set and the remainder is used as the test set.

In Table \ref{tab:dataset-stats}, we provide an overview of the different properties of these datasets. 

\begin{table*}[ht]
	\footnotesize
		\centering
	\caption{Overview of the datasets used to test the forecasting accuracy of the models.}
	\label{tab:dataset-stats}
	\resizebox{\columnwidth}{!}{\begin{tabular}{ccccccc}
		\toprule
		dataset  & domain & frequency & \# of time series & \# of timesteps & $T$ (context length) & $\tau$ (forecast horizon)\\
		\midrule
		 \exchange  & $\bR^+$ & daily & 8 & 6071 & 124 & 150\\
		\solar  & $\bR^+$ & hourly & 137 & 7009 & 336 & 168\\
		\electricity  & $\bR^+$ & hourly & 370 & 5790& 336 & 168\\
		\traffic  & $\bR^+$ & hourly & 963 & 10413 & 336 & 168\\
		\wiki  & $\mathbb{N}$ & daily & 2000 & 792 & 124 & 150\\
		\bottomrule
	\end{tabular}}
\end{table*}

\subsection{Forecasting metric: CRPS}
\label{app:crps}

The continuous ranked probability score (CRPS)~\cite{matheson1976scoring} is a proper scoring rule~\cite{gneiting2007strictly} that measures the compatibility of a quantile function $F^{-1}$ with an observation $y$.
It has an intuitive definition as the pinball loss integrated over all quantile levels  $\alpha \in [0, 1]$, i.e., 
\begin{equation}
\label{eq:crps}
\text{CRPS}(F^{-1},  y) = \int_0^1 2\Lambda_{\alpha}(F^{-1}(\alpha), y) \, d\alpha,
\end{equation}
where the pinball loss $\Lambda_{\alpha}(q, y) $ is defined as
\begin{equation}\label{eq:pinball}
\Lambda_{\alpha}(q, y) = (\alpha -\mathbbm{1}_{\{y < q\}})(y-q),
\end{equation}
with $q$ being the respective quantile of the probability distribution.

We used the CRPS implementation provided in GluonTS~\cite{alexandrov2020gluonts} that approximates the integral of \eqref{eq:crps} with a grid of selected quantiles.

\subsection{Training and hyperparameter details}
\label{app:forecast-hyperparam}

In this section we provide the training and hyperparameter details for the forecasting experiments.

\paragraph{Training parameters.} We trained all the datasets with a fixed batch size of 50 and tuned the number of training steps per dataset. We used the Adam optimizer for the gradient updates with $10^{-5}$ weight-decay and clipped the gradient norm to 10. The learning rate was warmed-up linearly from $1\times10^{-4}$ to a higher value in the range $[5\times10^{-4}, 1\times10^{-2}]$ (optimized per dataset) for the first $1000$ steps after which a cosine decay follows for the remaining time steps with a decay rate of 0.99.

\begin{table*}[htb]
	\footnotesize
	\centering
	\caption{\small Network architectures for the different components of RED-SDS for forecasting experiments. \texttt{Linear} denotes a linear layer without bias, \texttt{MLP} [$a_1$, \dots, $a_l$] denotes a $l$-hidden-layer MLP with hidden units $a_1$, \dots, $a_l$ and ReLU non-linearity, \texttt{biGRU} [$b$] denotes a single-layer bidirectional GRU with $b$ hidden units, \texttt{RNN} [$c$] denotes a single-layer RNN with $c$ hidden units, and \texttt{Transformer} denotes a transformer layer with one attention head and the dimension of feedforward network equal to 16.}
	\label{tab:fore_arch}
	\resizebox{\columnwidth}{!}{\begin{tabular}{lccccc}
		\toprule
		\multirow{2}{*}{Network} & \multicolumn{5}{c}{Datasets} \\
		\cmidrule{2-6}
		&      \exchange &        \solar & \electricity & \traffic & \wiki  \\
		\midrule
		Control Network ($f_u$) & \texttt{MLP} [32] & \texttt{MLP} [32] & \texttt{MLP} [32] & \texttt{MLP} [64] & \texttt{MLP} [32]\\
		Duration Network ($f_d$) & \texttt{MLP} [64] & \texttt{MLP} [64] & \texttt{MLP} [64] & \texttt{MLP} [64] & \texttt{MLP} [64]\\
		Discrete Transition ($f_z$) & \texttt{MLP} [$4 \times 2^2$] & \texttt{MLP} [$4 \times 4^2$] & \texttt{MLP} [$4 \times 3^2$] & \texttt{MLP} [$4 \times 4^2$] & \texttt{MLP} [$4 \times 5^2$]\\
		Continuous Transition ($f_x$) & \texttt{MLP} [32] & \texttt{Linear} & \texttt{MLP} [32] & \texttt{MLP} [32] & \texttt{Linear}\\
		Emission Network ($f_y$) & \texttt{MLP} [8, 32] & \texttt{Linear} & \texttt{MLP} [8, 32] & \texttt{MLP} [8, 32] & \texttt{Linear}\\
		\midrule
		Inference Embedder ($g^y_{\mathrm{emb}}$) & \texttt{biGRU} [4] & \texttt{Transformer}  & \texttt{Transformer} & \texttt{Transformer} & \texttt{Transformer}\\
		Causal RNN ($g_{\mathrm{rnn}}$) & \texttt{RNN} [16] & \texttt{RNN} [16] & \texttt{RNN} [16] & \texttt{RNN} [16] & \texttt{RNN} [16]\\
		Parameter Network ($g_{\mathrm{fc}}$) & \texttt{MLP} [32] & \texttt{MLP} [32] & \texttt{MLP} [32] & \texttt{MLP} [32] & \texttt{MLP} [32] \\
		\bottomrule
	\end{tabular}}
\end{table*}

\paragraph{Network types.} We experimented with linear and non-linear functions for the continuous transition $f_x$ and emission $f_y$ functions as previously described in Appendix \ref{app:seg-hyperparam}. For the inference embedding network $g^y_{\mathrm{emb}}$, we used a single  transformer layer with one attention head. The input time series was first mapped into a 4-dimensional embedding and then concatenated with the positional encoding before feeding into the transformer layer. For the control network ($f_u$) and the duration network ($f_d$) we used single hidden layer MLPs. Table \ref{tab:fore_arch} summarizes the network architectures of the different model components for the five forecasting datasets.

\begin{table*}[htb]
	\footnotesize
	\centering
	\caption{\small Control embedding hyperparameters.}
	\label{tab:fore_ctrl}
	\begin{tabular}{lccccc}
		\toprule
		\multirow{2}{*}{Hyperparameter} & \multicolumn{5}{c}{Datasets} \\
		\cmidrule{2-6}
		&      \exchange &        \solar & \electricity & \traffic & \wiki  \\
		\midrule
		$p$ & 5 & 8 & 32 & 50 & 8 \\
		$c$ & 16 & 16 & 32 & 32 & 32 \\
		\bottomrule
	\end{tabular}
\end{table*}

\paragraph{Control embedding.} As described in subsection \ref{subsec:instantiation}, the raw control features are comprised of static features $\vu^{\text{static}}$ and time features $\vu^{\text{time}}$. In our experiments, the static features are the time series IDs. Time series ID are first fed into an embedding layer that outputs a $p$-dimensional embedding; this time series embedding is concatenated with the time features and passed to the control network ($f_u$) to output a $c$-dimensional control $\vu_t$. Table \ref{tab:fore_ctrl} lists the values of $p$ and $c$ for the different datasets.

\paragraph{Duration.} We set $(\dmin, \dmax)$ to $(1, 20)$ for all datasets.

\paragraph{Normalization.} In many datasets considered by us, the scale of the time series varies significantly, sometimes by several orders of magnitude. This makes it difficult to train models---particularly those involving neural networks---and the individual time series require normalization before training. Such \emph{per time series} normalization is a standard practice for these datasets and has been employed in several previous works~\cite{salinas2020deepar,rangapuram2018deep,kurle2020deeprao}. To this end, we experimented with two normalization methods: standardization and scaling. Given a univariate time series $y_{1:T}$, the normalized version $y^{\text{norm}}_{1:T}$ for each method is computed as
\begin{align}
\text{Standardization:} \quad & y^{\text{norm}}_t = \frac{y_t - \text{mean}(y_{1:T})}{\text{std}(y_{1:T})},\\
\text{Scaling:} \quad & y^{\text{norm}}_t = \frac{y_t}{\frac{1}{T}\sum_{i=1}^T |y_i |},
\end{align}
where $\text{mean}(\cdot) $ and  $\text{std}(\cdot)$ denote the mean and the standard deviation, respectively. The normalization method for each dataset was tuned as a hyperparameter.

\paragraph{Log-determinant of the Jacobian.} 
Normalization of data is equivalent to applying a linear transformation to the raw input. More formally, consider a function $f$ that transforms a variable $\vv \in \mathbb{R}^D$ to the normalized variable $\vy \in \mathbb{R}^D$ via a function $f$, i.e., $f(\vv) = \vy$. If $p_V(\vv)$ is the probability density of the random variable $\vv$ and $f$ is an invertible and differentiable transformation, then by the change of variables formula the probability density $p_Y(\vy)$ of the transformed random variable $\vy$ is given by
\begin{equation}
\label{eq:change_var1}
\begin{aligned}
p_Y(\vy) & =  p_V(f^{-1}(\vy))  \left| \det \left( \frac{\partial f^{-1}(\vy)}{\partial \vy} \right) \right|, \\
\end{aligned}
\end{equation}
where $ \frac{\partial f^{-1}(\vy)}{\partial \vy} \in \mathbb{R}^{D\times D}$ is the Jacobian of $f^{-1}$. The determinant term accounts for the space distortion of the transformation, i.e., how the volume has changed locally due to the transformation.

Using the change of variables formula, this transformation corresponds naturally to a $\log\det$ term in the log-likelihood which is equal to $J = -\log{s}$, where $s=\text{std}(y_{1:T})$ in the case of standardization and $s=\sum_{i=1}^T |y_i |$ in the case of scaling (the Jacobian in both cases is a diagonal matrix).
In the forecasting experiments the $\log\det$ term is included in the objective during training.
Note that this holds only if we treat $s$ as a scaling factor that is independent of the time series $y_{1:T}$; although this is not true, in practice this training heuristic slightly improves the quantitative forecasting performance.

\paragraph{Number of switches.} In the case of segmentation, the ground truth number of switches $K$ is known (at least in the context of our experiments). For forecasting, we consider a range of $K\in \{2,\dots,5\}$ and we tuned it for each dataset.

\paragraph{Dimensionality of state $\vx_t$.} We tuned the dimensionality of the state variables $\vx_t$ from the set $\{2, 4, 8, 16\}$ for each dataset.

\subsection{Baseline models}
\label{app:forecast-baselines}

In this section, we provide additional details on the baseline forecasting models. The CRPS results for these baseline models have been taken from \cite{kurle2020deeprao}.

\textbf{ARSGLS}~\cite{kurle2020deeprao} is an extension of the vanilla SLDS that uses continuous (Gaussian) switch variables. It incorporates conditionally linear state-to-switch recurrence, keeping the conditional tractability of LDS, and augments the emission model by an auxiliary variable
that allows for modelling multivariate and non-Gaussian observations with complex non-linear transformations.
Inference in ARSGLS is performed via Rao-Blackwellised particle filters wherein the state conditional expectations are computed exactly using the available closed-form expressions whereas the switch and auxiliary variables expectations are approximated via Sequential Monte Carlo using a neural network.

The authors proposed two different instantiations of their model with differences in the underlying base LDSs. In the first instantiation, labelled as RSGLS-ISSM, the authors implemented the LDS as an innovation state space model (ISSM) with a constrained structure that models temporal patterns such as level, trend and seasonality where the transition and emission matrices are pre-defined and not learned. The second instantiation, labelled ARSGLS, uses an unconstrained LDS.

\textbf{KVAE}~\cite{fraccaro2017disentangled} uses a VAE to model auxiliary variables and a ``mixture'' of LDSs to model the dynamics. KVAE can be interpreted as an SLDS with deterministic switches, where the LDS parameters are predicted as a weighted average of a set of base matrices, with the weights given by an RNN. 
The RNN in the KVAE depends (autoregressively) on samples of the previous auxiliary variables.
Variational inference is performed using a recognition network for the auxiliary variables and Kalman filtering/smoothing for the LDSs' latent variables. 

The objective function in KVAE uses samples from the smoothing distribution. However, as noted in \cite{kurle2020deeprao}, the corresponding expectation can be computed in closed-form and the resulting objective function has lower variance. We report the results for both original KVAE (KVAE-MC) and the Rao-Blackwellised variant (KVAE-RB) proposed by \citet{kurle2020deeprao}.

\textbf{DeepState}~\cite{rangapuram2018deep} parametrizes an LDS using an RNN which is conditioned on inputs (controls).
The LDS parameters in DeepState have a fixed structure that model time series patterns such as level, trend and seasonality (same as in RSGLS-ISSM). 
The transition and emission matrices are fixed and the (diagonal) noise covariance matrices are predicted by the RNN directly.
Given the LDS parameters from the RNN, DeepState uses the Kalman filter for inference and maximum likelihood for parameter learning.

\textbf{DeepAR}~\cite{salinas2020deepar} is a strong \emph{discriminative} baseline model for probabilistic forecasting that uses an autoregressive RNN conditioned on the history of the time series (lags) and other relevant features. DeepAR autoregressively outputs the parameters of the future distributions and is trained using maximum likelihood estimation.

\section{Computational details}
\label{app:compute}

We used a \texttt{p3.8xlarge} AWS EC2 instance for running our experiments. This instance comprises 4 Tesla V100 GPUs, 36 CPUs, and 244 GB of memory.

\end{document}